\documentclass[12pt]{article}
\usepackage{mathtools}
\usepackage{amsmath,amssymb,amsfonts}
\usepackage{graphicx}
\usepackage{IEEEtrantools}
\usepackage{pgfplots}
\usepackage{booktabs, multirow, mleftright,bbm}
\usepackage[normalem]{ulem}
\usepackage{subfig}
\usepackage{times}
\usepackage{color}
\usepackage{multirow}
\usepackage[authoryear]{natbib}
\usepackage{setspace}

\usepackage[hidelinks]{hyperref}
\pgfplotsset{compat=newest}
\usepgfplotslibrary{groupplots, fillbetween}
\usetikzlibrary{positioning, fit, calc,shapes.misc,spy}

\newcommand{\sfootnote}[1]{\footnote{#1}}

\newcommand\blfootnote[1]{%
  \begingroup
  \renewcommand\thefootnote{}\footnote{#1}%
  \addtocounter{footnote}{-1}%
  \endgroup
}

\newcommand{\captionfonts}{}

\makeatletter  
\long\def\@makecaption#1#2{%
\vskip\abovecaptionskip
\sbox\@tempboxa{{\captionfonts #1: #2}}%
\ifdim \wd\@tempboxa >\hsize
{\captionfonts #1: #2\par}
\else
\hbox to\hsize{\hfil\box\@tempboxa\hfil}%
\fi
\vskip\belowcaptionskip}
\makeatother   

\newcommand{\badres}[1]{\color{gray}#1}
\newcommand{\goodres}[1]{\bfseries#1}

\newcommand{\figref}[1]{Fig.~\ref{#1}}
\newcommand{\figsref}[1]{Figs.~\ref{#1}}
\newcommand{\tabref}[1]{Table~\ref{#1}}
\newcommand{\tabsref}[1]{Tables~\ref{#1}}

\newcommand{\set}[1]{\mathcal{#1}} 

\newcommand{\argmin}{\operatornamewithlimits{argmin}}
\newcommand{\argmax}{\operatornamewithlimits{argmax}}

\newcommand{\Reals}{\mathbb{R}}

\newcommand{\T}{\mathsf{T}}

\newcommand{\Normal}[2]{\mathcal{N}\!\left({#1},{#2}\right)} 

\newcommand{\E}[1]{\mathbb{E}\!\left({#1}\right)} 

\newcommand{\taumin}{\tau_0}
\newcommand{\dmin}{d_\textrm{min}}
\newcommand{\dmax}{d_\textrm{max}}
\newcommand{\pr}{\operatorname{pr}}   
\newcommand{\rc}{\operatorname{rc}}   
\newcommand{\zmax}{\theta_{\mathrm{r}}}
\newcommand{\dzmin}{\dot{\theta}_{\mathrm{s}}}
\newcommand{\rhomax}{\rho_\textrm{max}}  

\begin{document}
\vspace{20mm}

{ \LARGE \noindent Continuous-Time Neural Networks Can Stably \\ Memorize Random Spike Trains \\}

\noindent 
{\bf \large Hugo Aguettaz and Hans-Andrea Loeliger}\\
{\{aguettaz, loeliger\}@isi.ee.ethz.ch}\\
{ETH Zürich, Dept. of Information Technology \& Electrical Engineering} \\


%
\ \vspace{-0mm}\\

\begin{center} {\bf Abstract} \end{center}
The paper explores the capability of continuous-time recurrent neural networks 
to store and recall 
precisely timed scores of spike trains.
We show (by numerical experiments) that this is indeed possible: 
within some range of parameters, any random score of spike trains 
(for all neurons in the network) 
can be robustly memorized and 
autonomously reproduced with stable accurate relative timing of all spikes, 
with probability close to one. 
We also demonstrate associative recall under noisy conditions.

In these experiments, the required synaptic weights are computed offline, 
to satisfy a template that encourages temporal stability.
\blfootnote{This manuscript is the authors' final version, accepted for publication in \emph{Neural Computation}. The final published version will be available from the journal.}

\section{Introduction}

Biological neural networks work in continuous time, without a clock, 
and with (apparently) imprecise and noisy neurons \citep{KandelAl21}.
It is thus not obvious how memories can be stably represented at all. 
Nonetheless, biological neural networks appear to have impressive storage capacity 
and the ability to control muscles 
with high temporal precision.
Understanding this better is of interest in its own right, 
and it may give ideas for designing future neuromorphic hardware 
\citep{IndiveriLiu15,KudithipudiAl25}.

A related different question 
is to what degree spiking neural networks 
can work with precisely timed spikes (rather than just with firing rates),
cf.\ \cite{VanRullenAl05}.
Polychronous groups of neurons have been observed 
in recurrent networks by \cite{Izhikevich06}
and more general patterns (motifs) with precisely timed firings
have been identified by \cite{GrimaldiAl23}.
In fact, most recent work with spiking neural networks 
relies (implicitly or explicitly) 
on precisely timed spikes, 
cf.\ \cite{PaugamMoisyBohte12,ComsaAl22} and also \cite{Neff2015}.
However, it is not clear if recurrent networks can autonomously 
operate with precisely timed spikes,
i.e., when the network is not kept in synchrony by and with external stimuli,
cf.\ \cite{BanerjeeAl08} and~\cite{LajoieAl13}.

In this paper, we consider an extreme version of memorization in recurrent networks, viz., 
the stable memorization of entire scores of spike trains as in \figref{fig:SpikeTrain}, 
with exact relative timing of all spikes of all neurons in the network. 
Our main result is the empirical observation that, with suitable qualifications, 
this is indeed possible:
in some range of parameters, for any random score of spike trains, 
there exist synaptic weights (with probability close to one) 
such that the network can autonomously reproduce all these spikes (with accurate relative timing), 
even in the presence of substantial disturbances.
To the best of our knowledge, this has not been demonstrated (and not even suggested) before.

\begin{figure}[t]
    \centering
    \begin{tikzpicture}[>=latex]
        \begin{groupplot}
            [
                group style={group size=1 by 4, vertical sep=2mm},
                width=0.8\textwidth,
                height=0.2\textwidth,
                tick pos=left,
                tick align=inside,
                xmin=-0.8, xmax=6,
                ymin=0, ymax=2.2, 
                xlabel=\empty,
                xtick={0,...,5},
                xticklabels=\empty,
                ytick=\empty, 
                axis x line=bottom, 
                x axis line style={->}, axis y line=left, 
                y axis line style={opacity=0}, 
                ylabel style={at={(axis description cs:0,0.4)}, anchor=east, rotate=-90},
                title style={at={(axis description cs:0.5,0.5)}, anchor=south},
            ]   

            \nextgroupplot[ylabel=$x_1(t)$]
            \addplot[ycomb,blue!80!black,mark=*,very thick] coordinates {(0.2,1.0) (1.6,1.0) (3.8,1.0) (4.9,1.0)};

            \nextgroupplot[ylabel=$x_2(t)$]
            \addplot[ycomb,blue!80!black,mark=*,very thick] coordinates {(1.5,1.0) (3.2,1.0)};
            \nextgroupplot[ytick=\empty, xticklabels=\empty, x axis line style={opacity=0}, xtick=\empty, height=0.32*\axisdefaultheight]
            \node at (axis cs:-0.4,0.8) {$\vdots$};
            \node at (axis cs:2.4,0.8) {$\vdots$};
            \node at (axis cs:5.2,0.8) {$\vdots$};
            \nextgroupplot[ylabel=$x_L(t)$, xlabel={$t$}]
            \addplot[ycomb,blue!80!black,mark=*,very thick] coordinates {(0.8,1.0) (3.2,1.0) (5.1,1.0)};
        \end{groupplot}
    \end{tikzpicture}
    \caption{\label{fig:SpikeTrain}%
    A score of spike trains $x_1, ..., x_L$ produced by $L$ neurons, 
    with spikes (=~\mbox{firings})
    of the same neuron separated by at least $\taumin$ (=~distance between the tickmarks).
    The paper is about memorizing and reproducing such scores.
    }
\end{figure}

Our experiments also demonstrate 
that clockless continuous-time networks can operate 
with global spike-level temporal stability almost like digital processors,
which does seem to open interesting perspectives for neuromorphic engineering.

For technical convenience, we will formulate our memorization problem 
in terms of cyclic attractors.
However, our perspective is quite different from the classical view of cyclic
attractors as individual elements of memory 
\citep{Amit89,PoucetSave05}.
Instead, we mostly consider the whole memory as one single cyclic attractor,
i.e., the entire memory content consists of one single long periodic score of spike trains
as in Fig.~\ref{fig:SpikeTrain}. 
(Partitioning the memory into several cyclic attractors
is briefly addressed in Sections \ref{sec:Capacity} and ~\ref{sec:AssociativeRecall}.)

We will work with (a variation of) a standard continuous-time neural network model 
with random transmission delays.
The latter 
are essential;
we thus reconfirm the observation from \citep{Izhikevich06, MarziMadhow2018, Perrinet23, DeckersAl24}
that heterogeneous axonal delays are not a nuisance, but a valuable resource.
These transmission delays are fixed, which differs from some recent work
where both the synaptic weights and the transmission delays are learned 
\citep{Perrinet23, HammouamriAl23, DeckersAl24}.

A novel feature of our neural network model is threshold noise (cf.\ Section~\ref{sec:Threshold}),
which models the inherent noisiness and imprecision of neural computation 
(both biological and with neuromorphic hardware) in a simulation friendly way.

The computation of the synaptic weights uses an optimization algorithm
with a new condition for encouraging temporal stability.
While we do not claim biological plausibility for this algorithm, it is at least neuron local, 
i.e., the synaptic weights to each neuron are computed using only (spike time) information 
that this neuron sees in its own operation.

The paper is structured as follows. 
The network model is defined in Section~\ref{sec:Network}.
The precise problem statement 
and the primary experimental results are given in Section~\ref{sec:ProblemStatement}.
The rest of the paper explores the role of some parameters 
and elaborates on technical details.
The computation of the synaptic weights 
is described and discussed 
in Section~\ref{sec:compute_weights} and \ref{app:Optimization}.
The capacity of such networks is discussed in Section~\ref{sec:Capacity}.
Associative recall is demonstrated in Section~\ref{sec:AssociativeRecall},
and the conclusions are offered in Section~\ref{sec:Conclusion}.

Many technical details are given in the appendices.
The software for the experiments in this paper 
is available on GitHub%
\sfootnote{\url{https://github.com/haguettaz/rusty-snn}}.

\section{Spiking Neural Network Model} \label{sec:Network}

\subsection{Spike Trains and Network Structure} \label{sec:SpikeNetwork}

We will use recurrent networks with the structure of \figref{fig:Network}.
There are $L$ neurons, producing spike trains $x_1(t),\ldots,x_L(t)$ as illustrated in \figref{fig:SpikeTrain}. 
Every neuron has $K$ inputs, and each of these inputs is randomly selected (independently and with uniform probability) from $x_1(t),\ldots,x_L(t)$.
The inputs of neuron $\ell$ are denoted 
$\tilde x_{\ell,1}, \ldots, \tilde x_{\ell,K}$.
Note that both parallel connections (between the same pair of neurons) 
and self-connections (some neuron feeding itself) are allowed.
Note also that $K$ and $L$ are unrelated: both $K\ll L$ and $K\gg L$ are possible.

Each connection has its own 
axonal delay, 
which, however, will be viewed as part of the neuron model,
which will be specified below.

For mathematical convenience, the spikes are 
represented by 
Dirac deltas.
Thus each spike train $x_\ell(t)$ is of the form
\begin{equation} \label{eqn:SpikeTrain}
    x_\ell(t) = \sum_{s \in \set{S}_\ell} \delta(t - s),
\end{equation}
where $\set{S}_\ell \subset\Reals$ is the (discrete) set of firing times of neuron $\ell$.
The neuron model (specified below) guarantees that any two spikes of the same neuron are separated by at least $\taumin > 0$,
which is the natural unit of time in the network.

\begin{figure}[t]
    \centering
    \begin{tikzpicture}[scale=0.1,>=latex]
        \node [draw, rectangle, dashed, anchor=south west, minimum width = 15mm, minimum height = 55mm] 
              (block) at (10,2.5)
              {\rotatebox{90}{random connections}};
        
        \draw (47.5,31) node{$\vdots$};
        \draw (5,31) node{$\vdots$};
                
        \draw (0,65) -- (0,46);
        \draw[->] (0,54) -- (10,54);
        \filldraw (0,54) circle(0.5);
        \draw (5,51) node{$\vdots$};
        \draw[->] (0,46) -- (10,46);
        
        \draw[->] (25,54) -- node[above]{$\tilde x_{1,1}$} (40,54);
        \draw (32.5,51) node{$\vdots$};
        \draw[->] (25,46) -- node[below]{$\tilde x_{1,K}$} (40,46);
        
        \node [draw, rectangle, minimum width = 20mm, minimum height = 12mm, anchor=west]
              (neuron1) at (40,50)
              {neuron 1};
        
        \draw (neuron1) -- node[below]{$x_1$} +(17.5,0) -- +(17.5,15) -- (0,65);

        \draw[->] (0,14) -- (10,14);
        \draw (5,11) node{$\vdots$};
        \draw[->] (0,6) -- (10,6);
        \filldraw (0,6) circle(0.5);
        \draw (0,-5) -- (0,14);
        
        \draw[->] (25,14) -- node[above]{$\tilde x_{L,1}$} (40,14);
        \draw (32.5,11) node{$\vdots$};
        \draw[->] (25,6) -- node[below]{$\tilde x_{L,K}$} (40,6);
        
        \node [draw, rectangle, minimum width = 20mm, minimum height = 12mm, anchor=west]
              (neuronL) at (40,10)
              {neuron $L$};
        
        \draw (neuronL) -- node[above]{$x_L$} +(17.5,0) -- +(17.5,-15) -- (0,-5);
        
        \end{tikzpicture}
    \caption{Network structure with $L$ neurons (producing spike trains $x_1,\ldots, x_L$) connected at random. 
    Every neuron receives the same number $K$ of inputs.}
    \label{fig:Network}
\end{figure}

\subsection{The Spiking-Neuron Model} \label{sec:Neuron}

We will use a version of the Spike Response Model (SRM) \citep{Gerstner95, KistlerAl97},
which offers
just enough biological (or physical)
realism to be useful for our purpose.

\begin{figure}[t]
\centering
\begin{tikzpicture}[>=latex,node distance=6mm and 8mm]
    \tikzset{midbox/.style={draw, rectangle, minimum size=10mm, inner sep=4pt}}
    \tikzset{bigbox/.style={draw, rectangle, minimum height=14mm, minimum width=18mm, inner sep=4pt}}
    \tikzset{mult/.style={draw, circle, minimum size=12mm}}
    \tikzset{adder/.style={draw, circle, minimum size=6mm, inner sep=1pt}}
    \tikzset{dot/.style={draw, circle, fill=black, minimum size=2pt, inner sep=0pt}}

    \node[adder] (adder) {$+$};

    \node[mult, above left=8mm and 4mm of adder, label={center:$w_{\ell,1}$}] (w1) {};
    \node[mult, below left=8mm and 4mm of adder, label={center:$w_{\ell,K}$}] (wK) {};

    \node[bigbox, left=of w1, text width=10mm, align=center] (delay1) {delay $d_{\ell,1}$};
    \node[bigbox, left=of wK, text width=10mm, align=center] (delayK) {delay $d_{\ell,K}$};
    \node at ($(delay1)!0.48!(delayK)$) {$\vdots$};

    \node[bigbox, right=6mm of adder] (h) {};

    \begin{axis}[at=(h), anchor=center, width=28mm, height=22mm, xmin=0, xmax=8, ymin=0, ymax=1.1, clip=false, axis lines=none]
        \addplot[thick, blue!60!black, domain=0:8, smooth] plot (\x,{\x*exp(1-\x)});
    \end{axis}

    \node[bigbox, right=8mm of h, align=center, text width=16mm] (fire) {fire if $z_\ell \geq \theta_{\ell}$};

    \draw[<-] (delay1.west) -- ++(-0.8,0) node[above, pos=0.55]{$\tilde x_{\ell,1}$};
    \draw[->] (delay1) -- (w1);
    \draw[->] (w1) -| (adder);
    \draw[->] (delayK) -- (wK);
    \draw[<-] (delayK.west) -- ++(-0.8,0) node[above, pos=0.55]{$\tilde x_{\ell,K}$};
    \draw[->] (wK) -| (adder);
    \draw[->] (adder) -- (h);
    \draw[->] (h) -- (fire) node[above, midway]{$z_\ell$};
    \draw[->] (fire.east) -- ++ (0.8,0) node[above, midway]{$x_{\ell}$};
\end{tikzpicture}   
\caption{\label{fig:Neuron}%
Neuron model with potential \eqref{eqn:PotentialSum}.}
\end{figure}

The pivotal quantity of each neuron is its potential $z_\ell(t)$: 
the neuron fires (i.e., it produces a spike) whenever $z_\ell(t)$ crosses a threshold $\theta_\ell(t)$, 
as will be detailed in Section~\ref{sec:Threshold}.

As illustrated 
in \figref{fig:Neuron}, $z_\ell(t)$ develops according to
\begin{equation} \label{eqn:PotentialSum}
    z_{\ell}(t) = \sum_{k=1}^K w_{\ell,k}
    \sum_{s \in \tilde{\mathcal{S}}_{\ell,k}}
    h(t - s -  d_{\ell,k})
\end{equation}
where $\tilde{\mathcal{S}}_{\ell,k}$ is the (discrete) set of firing times of $\tilde x_{\ell,k}$ and 
\begin{equation} \label{eqn:ImpulseResponse}
    h(t) \triangleq
    \begin{dcases}
        \frac{t}{\beta}
        \exp \mleft( 1 - \frac{t}{\beta} \mright) & \text{if } t > 0 \\
        0 & \text{if } t \leq 0.
    \end{dcases}
\end{equation}

It is easily seen that the pulse (\ref{eqn:ImpulseResponse}) peaks at
$t = \beta$
with $\max_t h(t) = 1$.
Therefore, the contribution of any single spike in $\tilde x_{\ell,k}$ to (\ref{eqn:PotentialSum}) is bounded in magnitude by $|w_{\ell, k}|$.

The exact pulse shape (\ref{eqn:ImpulseResponse}) is not essential: 
any similar pulse shape would do (i.e., yield qualitatively similar results); 
however, the specific shape (\ref{eqn:ImpulseResponse}) is both natural and
allows for fast network simulation, 
cf.\ \cite{ComsaAl22} and \ref{app:NetworkSimulation}.

The individual axonal delays $d_{\ell, k}$ are chosen at random (independently and uniformly between $d_\text{min}$ and $d_\text{max}$), but fixed during operation. 
These delays are essential for this paper.

The weights $w_{\ell,k}\in\Reals$ are available for learning.
We will always use bounded weights.

\subsection{Refractory Period and Threshold Noise}
\label{sec:Threshold}

A neuron fires when its potential (\ref{eqn:PotentialSum}) 
exceeds a threshold~$\theta_\ell$, 
after which the neuron cannot fire for a period of duration~$\taumin$. 

All neurons have the same fixed nominal threshold $\theta_0>0$.
However, 
a realistic model of biological neural networks or neuromorphic hardware 
should allow for noise in the computations that manifests in randomly shifted, 
false, or missed firings \citep{FaisalAl08}.
In this paper, 
we model such noise by piecewise constant random thresholds $\theta_\ell(t)$, as follows:
after each firing of neuron $\ell$,
its threshold $\theta_\ell$ (for the next firing)
is newly (and independently) 
sampled from a Gaussian distribution with mean $\theta_0$ and variance $\sigma_\theta^2$, 
as illustrated in \figref{fig:Threshold}.

Note that this noise model is easy to simulate, has only one parameter ($\sigma_\theta^2$), 
but captures the essential effects: 
small deviations from $\theta_0$ result primarily in shifted firings
while large deviations from $\theta_0$ can suppress or produce firings.

\begin{figure}[t]
    \centering
    \begin{tikzpicture}[>=latex]
        \begin{axis}
            [
                width=0.9\textwidth,
                height=0.3\textwidth,
                tick pos=left,
                tick align=inside,
                xmin=-2.0,xmax=10.5,
                ymin=0,ymax=2,
                xlabel=\empty,
                xtick={0, 2.2, 8.1},
                xticklabels={$s_{n-1}$, $s_n$, $s_{n+1}$},
                ytick={1},
                yticklabels={$\theta_0$},
                clip=false,
                grid=major,
                xmajorgrids=false
            ]   

            \draw[blue!80!black, thick] (-2,0.6) -- (0,0.6);
            \draw[blue!80!black, thick] (0,1.3) -- (2.2,1.3);
            \draw[blue!80!black, thick] (2.2,0.9) -- (8.1,0.9);
            \draw[blue!80!black, thick] (8.1,1.5) -- (10.5,1.5);

            \fill[blue!80!black] (0,0.6) circle (2pt);
            \fill[blue!80!black] (2.2,1.3) circle (2pt);
            \fill[blue!80!black] (8.1,0.9) circle (2pt);
        \end{axis}
    \end{tikzpicture}
    \vspace{-0.5cm}
    \caption{Noisy threshold $\theta_\ell(t)$ with firings at times $s_{n-1}$, $s_n$, $s_{n+1}$.}%
    \label{fig:Threshold}
\end{figure}

\section{Problem Statement and Main Result} 
\label{sec:ProblemStatement}

Suppose we are given 
\begin{itemize}
\item[(a)]
a network as described in Section~\ref{sec:Network}, and
\item[(b)]
a prescribed score of spike trains 
$\breve x = (\breve x_1, \breve x_2, ..., \breve x_L)$ as in \figref{fig:SpikeTrain},
with firings of the same neuron separated by at least $\taumin$.
\end{itemize}
Do there exist weights $w_{\ell,k}$ (for all neurons)
such that the network (when properly initialized) 
can autonomously (i.e., without an external input) and stably
(even with threshold noise)
reproduce the prescribed firing score?

The main result of this paper is 
that such weights indeed exist
(for some range of parameters, with some qualifications), 
and that we can actually compute them.
However, the problem statement itself requires
a number of clarifications.

\subsection{Just Standard Supervised Learning?}

Obviously, the stated problem may be viewed as a form 
of supervised learning, where each neuron is trained 
to produce
its next prescribed firing.
However, standard methods of supervised learning \citep{WangAl20}
will not enable the network to 
stably reproduce the prescribed score of spike trains%
\sfootnote{%
This issue does not arise in supervised training of feedforward networks 
as, e.g., in \citep{TapsonAl13,LeeAl17,ComsaAl22}
where the network is kept in synchrony by and with externally supplied input spike trains. 
}
autonomously:
small errors in the firing times will quickly accumulate
and the temporal coherence of the spike score will be lost
(cf.\ the first row of \tabref{tab:TemplateCondSlope}, 
which will be discussed in Section~\ref{sec:Capacity}).

Therefore, in this paper, 
it is critical to compute 
the synaptic weights in a way that 
encourages temporal stability.

\subsection{The Prescribed Score of Spikes: Periodic and Random}
\label{sec:PrescribedScore}

In order to simplify the discussion,
the prescribed score of spike trains is periodic
with period $T \gg \taumin$.
In other words, we wish the memorized spike score to be a cyclic attractor.
(The generalization to multiple cyclic attractors is briefly
addressed in Sections \ref{sec:Capacity} and ~\ref{sec:AssociativeRecall}.)

We will consider \emph{random} periodic spike scores,
which we wish the network to reproduce with probability close to one.
In fact, it is easy to see that there exist spike scores 
that the network cannot be made to reproduce%
\sfootnote{E.g., a neuron cannot fire if it receives no firings at any of its synapses.
Or some neuron sees the same past inputs twice, but is supposed to fire only in one of the two cases.},
but we wish the probability of all such unfeasible spike scores to be close to zero. 

The $L$ random spike trains (one for each neuron) are generated independently.
The probability law of each spike train is a suitably modified Poisson process with spikes separated by at least $\taumin$.
Both the exact definition and the practical generation (i.e., the sampling) of such random spike trains are 
detailed in \ref{app:SpikeTrain}.
In particular, the expected number of spikes (of each neuron, per period $T$) 
as a function of the firing rate $\lambda$ is 
given by \eqref{app:eqn:ExpNumSpikes} and plotted in \figref{fig:PMFNumSpikes}.

\subsection{Measuring Accuracy and Stability}
\label{sec:AccuracyMeas}

\begin{figure}[t]
    \centering
    \begin{tikzpicture}[>=latex]
        \begin{groupplot}
            [
                group style={group size=1 by 2, vertical sep=2mm},
                width=0.7\textwidth,
                height=0.2\textwidth,
                tick pos=left,
                tick align=inside,
                xmin=-0.2, xmax=6.5,
                ymin=0, ymax=2.2, 
                xlabel=\empty,
                xtick={0,...,6},
                ylabel=\empty,
                xticklabels=\empty,
                ytick=\empty, axis x line=bottom, x axis line style={->}, axis y line=left, y axis line style={opacity=0}, ylabel style={at={(axis description cs:0,0.3)}, anchor=east, rotate=-90},
                title style={at={(axis description cs:0.5,0.5)}, anchor=south},
            ]   

            \nextgroupplot[ylabel=$\breve x_\ell(t)$]
            \addplot[ycomb,blue!80!black,mark=*,very thick] coordinates {(0.8,1.0) (3.2,1.0) (4.3,1.0) (5.8,1.0)};

            \nextgroupplot[ylabel=$(\kappa \ast \breve x_\ell)(t)$, xlabel={$t$}]
            \addplot[blue!80!black, thick] coordinates { 
                 (-0.4,0) (0.3,0.0) (0.8,1.0) (1.3,0.0) (2.7,0.0) (3.2,1.0) (3.7,0.0) (3.8,0.0) (4.3,1.0) (4.8,0.0) (5.3,0.0) (5.8,1.0) (6.3,0.0) (6.5,0.0)};
        \end{groupplot}
    \end{tikzpicture}
    \caption{The ($T$-periodic) signal $\kappa * \breve x_\ell$ 
    in \eqref{eqn:Precision} and \eqref{eqn:Recall}. The distance between the tickmarks is $\taumin$.}
    \label{fig:KernelConv}
\vspace{\floatsep}

    \centering
    \begin{tikzpicture}[>=latex]
        \begin{groupplot}
            [
                group style={group size=1 by 3, vertical sep=3mm},
                width=0.8\textwidth,
                height=0.2\textwidth,
                tick pos=left,
                tick align=inside,
                xmin=-0.5, xmax=7.5,
                ymin=0, ymax=2.2, 
                xlabel=\empty,
                xtick={0,...,7},
                ylabel=\empty,
                xticklabels=\empty,
                ytick=\empty, axis x line=bottom, x axis line style={->}, axis y line=left, y axis line style={opacity=0}, 
                ylabel style={at={(axis description cs:0,0.35)}, anchor=east, rotate=-90},
                title style={at={(axis description cs:0.5,0.5)}, anchor=south},
            ]   

            \nextgroupplot[ylabel=$x_1(t)$]
            \addplot[ycomb,blue!80!black,mark=*,very thick] coordinates {(0.2,1.0) (1.6,1.0) (3.8,1.0) (4.9,1.0)};
            \addplot[ycomb,gray,mark=*,very thick] coordinates {(6.4,1.0) };
            \draw[dashed, blue!80!black] (0,-1) rectangle (6,1.5);
            \node[blue!80!black, anchor=north east] at (6,1.5) {$\mathcal{S}^{t_0}_1$};

            \nextgroupplot[ylabel=$x_2(t)$]
            \addplot[ycomb,blue!80!black,mark=*,very thick] coordinates {(1.5,1.0) (3.2,1.0) (6.1,1.0)};
            \addplot[ycomb,gray,mark=*,very thick] coordinates {(-0.2, 1.0)};
            \draw[dashed, blue!80!black] (0,-1) rectangle (7,1.5);
            \node[blue!80!black, anchor=north east] at (7,1.5) {$\mathcal{S}^{t_0}_2$};

            \nextgroupplot[ylabel=$x_3(t)$, xlabel={$t$},xticklabels={$t_{0}$,,,,,,$t_0 + T$}]
            \addplot[ycomb,blue!80!black,mark=*,very thick] coordinates {(0.2,1.0) (3.2,1.0)};
            \addplot[ycomb,gray,mark=*,very thick] coordinates {(5.7,1.0)};
            \draw[dashed, blue!80!black] (0,-1) rectangle (5,1.5);
            \node[blue!80!black, anchor=north east] at (5,1.5) {$\mathcal{S}^{t_0}_3$};
        \end{groupplot}
    \end{tikzpicture}
    \caption{\label{fig:Window}%
    A technical detail:
    the window $[t_0, t_0+T+c) \approx [t_0, t_0+T)$ in the definition of $S_\ell^{t_0}$ 
    in \eqref{eqn:Precision} and \eqref{eqn:Recall}. 
    The distance between the tickmarks is $\taumin$.}
\end{figure}

In our numerical experiments, the accuracy of reproduction
over some period $[t_0, t_0+T)$ 
will be measured by 
the precision
\begin{equation} \label{eqn:Precision}
    \pr(t_0)
    \triangleq 
    \max_{0 \leq \tau < T} \frac{1}{L} 
    \sum_{\ell=1}^L \frac{1}{|\set{S}_\ell^{t_0}|} \sum_{s \in \set{S}_\ell^{t_0} } (\kappa * \breve x_\ell)(s - \tau)
\end{equation}
if $|\set{S}_\ell^{t_0}|>0$ and $\pr(t_0) \triangleq  0$ otherwise,
and the recall
\begin{equation}  \label{eqn:Recall}
    \rc(t_0)
    \triangleq 
    \max_{0 \leq \tau < T} \frac{1}{L} 
    \sum_{\ell=1}^L \frac{1}{|\breve{\set{S}}_\ell^0|}\sum_{s \in \set{S}_\ell^{t_0}} (\kappa * \breve x_\ell)(s - \tau)
\end{equation}
if $|\breve{\set{S}}_\ell^{0}|>0$ and $\rc(t_0) \triangleq  0$ otherwise,
where 
\begin{itemize}
    \item
$\breve{x}_\ell$ 
is the periodic spike train prescribed for neuron~$\ell$ 
and $\breve{\set{S}}_\ell^0$ 
is the set of firing times within $[0,T)$,
\item
$x_\ell$ 
is the actual spike train produced by neuron~$\ell$ 
and $\set{S}_\ell^{t_0}$ 
is the set of firing times within $[t_0, t_0 + T + c) \approx [t_0, t_0+T)$,
where $c$ adjusts for border effects%
\sfootnote{%
Specifically, 
$c$ is the largest element of $\{-\taumin, 0, \taumin \}$
such that $\min_k |s + kT - s'| \geq \taumin$ for all $s \neq s' \in \set{S}_\ell^{t_0}$,
see \figref{fig:Window}.
}
that can arise since the actual firings are not strictly periodic,
\item
$|\breve{\set{S}}_\ell^0|$ and $|\set{S}_\ell^{t_0}|$ 
denote the cardinality of $\breve{\set{S}}_\ell^0$ and $\set{S}_\ell^{t_0}$,
respectively (i.e., the number of spikes in these sets),
\item 
\begin{equation} \label{eqn:Kernel}
    \kappa(t) \triangleq
    \begin{cases}
        1 - 2 |t|/\taumin , & |t| \leq \taumin/2 \\
        0, & \text{otherwise}
    \end{cases}
\end{equation}
is a triangular kernel
that covers at most one spike%
\sfootnote{%
By contrast, the SpikeShip kernel from \cite{VinckAl23} 
extends over the whole spike train.
}
(cf.\ \figref{fig:KernelConv}).
\end{itemize}
The maximizations in \eqref{eqn:Precision} and \eqref{eqn:Recall} 
aim to determine the best temporal adjustment 
between $x$ and $\breve x$ over one period.

Note that $0\leq \pr(t_0) \leq 1$,
and $\pr(t_0) = 1$ if and only if, for every $\ell$,
every spike in $\set{S}_\ell^{t_0}$ is perfectly matched by a spike in $\breve{\set{S}}_\ell^0$.
Likewise, $0\leq \rc(t_0) \leq 1$,
and $\rc(t_0) = 1$ if and only if, for every $\ell$,
every spike in $\breve{\set{S}}_\ell^0$ is perfectly matched by a spike in $\set{S}_\ell^{t_0}$.

Clearly, \eqref{eqn:Precision} and \eqref{eqn:Recall} 
coincide if $|\set{S}_\ell^{t_0}| = |\breve{\set{S}}_\ell^0|$;
in this case, small values of $\pr(t_0)$ and $\rc(t_0)$ 
indicate strong firing jitter.
The condition
$\pr(t_0) \gg \rc(t_0)$ 
indicates many missing firings
while $\pr(t_0) \ll \rc(t_0)$ 
indicates many false firings.

The condition
\begin{equation} \label{eqn:AccuracyThresholdCondition}
\pr(t_0)>0.9 \text{~~~and~~~} \rc(t_0)>0.9
\end{equation}
for $t_0 \gg T$
may be used as a (conservative) indicator for correct and stable memorization.

\subsection{Drift and Very-Long-Term Stability}
\label{sec:VeryLongTermStability}

Any two physical clocks that are not synchronized will inevitably drift 
away from each other (unless there is absolutely no noise).
This applies also
to the problem statement in this section:
the threshold noise may have a net effect of 
pushing all spikes forward (or backward) in time, which is unrecoverable
(see also Section~\ref{sec:EigenStability}).
Therefore, our measures of precision (\ref{eqn:Precision}) and recall (\ref{eqn:Recall})
cover only a single period rather than the whole time axis.

Moreover, 
the reader may have noticed that, with threshold noise as in Section~\ref{sec:Threshold},
the condition~(\ref{eqn:AccuracyThresholdCondition}) cannot be maintained forever.
For example, it is possible that the momentary thresholds $\theta_\ell(t)$
(i.e., $\theta_0$ plus threshold noise) 
of all neurons are set to such high values that all firings are suppressed,
which stalls the network in that state.
However, in some range of parameters,  
the probability of such derailments is so small 
that they are never observed in numerical experiments.

\subsection{Analytical Stability} 
\label{sec:EigenStability}

A necessary condition for temporal stability 
is that the propagation of spike jitter through the network
is damped rather than increased. 
For small jitter, this can be analyzed by the eigenvalues of a linearized model,
as was done in \citep{BanerjeeAl08}. 

Let $(\breve s_{n})_{n \in \mathbb{N}}$
be a sequence of (sorted) nominal spike positions of all spikes,
(i.e., the prescribed firing times of all neurons together)
and let $s_n \triangleq \breve s_n+ds_n$, $n \in \mathbb{N}$, be the corresponding actual spike positions.
Assume that the prescribed firing score is periodic 
and let $N$ be the total number of spikes in the first period. 
Assume that there is no threshold noise 
and $|ds_n| \ll \taumin$ 
for all $n \in \{ 0, 1, ..., N-1\}$.
For $n \geq N$,
the timing errors $ds_n$, as long as their magnitude is small, 
propagate according to
\begin{equation} \label{eqn:LinJitterProp}
ds_n \approx \sum_{n'=1}^N  a_{n,n'} \, ds_{n-n'}
\end{equation}
with
\begin{equation} \label{eqn:LinJitterPropCoeffs}
a_{n,n'} \triangleq \frac{\partial s_n}{\partial s_{n-n'}}
\end{equation}
where we assume (without loss of essential generality)
that the influence of $ds_{n-n'}$ for $n'>N$ can be neglected.
The derivatives \eqref{eqn:LinJitterPropCoeffs}
are given by \eqref{app:eqn:LinJitterPropCoeffs} in \ref{app:LinJitterProp}.
Note that $a_{n,n'} = a_{n+N,n'}$ for every $n,n'$.

For $n > 0$,
the jitter vector 
$\Delta_n \triangleq (ds_{n + N-1}, \ldots, ds_{n})^\T$
thus (approximately) evolves according to the linear recurrence 
\begin{equation} \label{eqn:JitterRec}
    \Delta_n = A_n \Delta_{n-1}
\end{equation}
with
\begin{equation} \label{eqn:JitterA}
    A_n \triangleq 
    \begin{pmatrix}
        a_{n,1} & \ldots & a_{n,N-1} & a_{n,N} \\
        & I_{N-1} & & 0_{N-1}
    \end{pmatrix}
    \in
    \Reals^{N\times N},
\end{equation}
where $I_{N-1}$ is an $(N-1)\times (N-1)$ identity matrix
and $0_{N-1}$ is an all-zeros column vector of size $N-1$.
By recursion, \eqref{eqn:JitterRec} yields
\begin{equation} \label{eqn:JitterRecExp}
    \Delta_n = \Phi_{n} \Delta_0 
\end{equation}
with 
$\Phi_n \triangleq A_n A_{n-1} \cdots A_{1}$
(for $n > 0$).
Since the sequence $A_n, A_{n+1},\ldots$ of matrices
is periodic with period $N$, 
\eqref{eqn:JitterRecExp} can be written as
\begin{equation}
    \Delta_{n} = \Phi_{r} \Phi^{q}_{N} \Delta_0,
\end{equation}
where $q$ and $r$ are the quotient and the remainder, respectively, 
of the division of $n$ by $N$.
The behavior of  $\Delta_{n}$ for $n \to \infty$ 
is thus governed by the 
dominant eigenspace of $\Phi_{N}$.

The rows of the matrix $A_n$ are easily seen to sum to one.
It follows that the rows of $\Phi_n$ also sum to one,
which implies that it has an all-ones eigenvector with eigenvalue~1,
corresponding to a uniform shift of all spikes.
Since a uniform shift of all spikes is irrelevant for temporal stability 
(cf.\ Section~\ref{sec:VeryLongTermStability}), 
we can get rid of this eigenvector 
by deflating $\Phi_{N}$ to 
\begin{equation} \label{eqn:Deflated}
    \tilde \Phi_{N} \triangleq \Phi_{N} - \frac{1}{N} J_{N},
\end{equation}
where $J_{N}$ is the all-ones matrix of size $N \times N$;
the all-ones vector is in the kernel of the matrix (\ref{eqn:Deflated}),
which otherwise has the same eigenpairs as $\Phi_{N}$ \citep{saad2011}.

Let 
$\rhomax$
be the spectral radius of $\tilde \Phi_{N}$,
i.e., the maximum of the magnitudes of the eigenvalues of $\tilde \Phi_{N}$.
The jitter vector $\Delta_n$ will die out for $n \to \infty$
if and only if 
\begin{equation} \label{eqn:SpecRadCond}
    \rhomax < 1;
\end{equation}
for $\rhomax > 1$,  $\Delta_n$ will grow exponentially with~$n$.

The condition \eqref{eqn:SpecRadCond}
is necessary, but not sufficient,
for temporal stability 
with nonvanishing threshold noise
(as measured by \eqref{eqn:AccuracyThresholdCondition}). 
However, computing $\rhomax$
is less time-consuming than, and provides complementary insight to,
simulations of the network.

\subsection{Main Result}
\label{sec:ThresholdNoiseStability:Results}

\begin{table}[tp]
    \centering
    \caption{\label{tab:ThresholdNoiseStabilityNeurons}%
    Experimental demonstration of accurate stable reproduction,
    measured by the precision~(\ref{eqn:Precision}) and the recall~(\ref{eqn:Recall}),
    for different levels of threshold noise (parameterized by $\sigma_{\theta}$).
    Each row summarizes 10 repetitions of the same experiment (minimum, median, and maximum).
    Also shown is the logarithm of the spectral radius (\ref{eqn:SpecRadCond}), 
    which does not depend on $\sigma_{\theta}$.
    Bold type indicates success 
    (i.e., precision and recall above the thresholds in (\ref{eqn:AccuracyThresholdCondition})), 
    faint gray indicates failure.
    }
    \vspace{0.5em}
    \setlength{\tabcolsep}{3pt}
    \begin{tabular}{c@{\hspace{1.2em}}c@{\hspace{1em}}ccc@{\hspace{1em}}ccc@{\hspace{1.5em}}cc}
        \toprule
        & 
        & \multicolumn{3}{c}{$\pr(50T)$} 
        & \multicolumn{3}{c}{$\rc(50T)$}
        & \multicolumn{2}{c}{$\ln \rhomax$} \\
        $L$ 
        & $\sigma_{\theta} / \theta_0$ 
        & $\min$ 
        & $\mathrm{med}$ 
        & $\max$ 
        & $\min$ 
        & $\mathrm{med}$ 
        & $\max$ 
        & $\min$ 
        & $\max$ 
        \\
        \midrule
        & 5\%
        & \goodres{0.978} 
        & \goodres{0.979} 
        & \goodres{0.980} 
        & \goodres{0.978} 
        & \goodres{0.979} 
        & \goodres{0.980} 
        \\
        50 
        & 10\% 
        & \goodres{0.953} 
        & \goodres{0.957} 
        & \goodres{0.960} 
        & \goodres{0.953} 
        & \goodres{0.957} 
        & \goodres{0.960} 
        & \goodres{-- 7.2}
        & \goodres{-- 6.2}
        \\
        & 20\% 
        & \badres{0.151} 
        & \badres{0.172} 
        & \badres{0.195} 
        & \badres{0.126} 
        & \badres{0.150} 
        & \badres{0.176} 
        \\
        \midrule 
        & 5\% 
        & \goodres{0.978} 
        & \goodres{0.979} 
        & \goodres{0.980} 
        & \goodres{0.978} 
        & \goodres{0.979} 
        & \goodres{0.980} 
        \\
        100 
        & 10\% 
        & \goodres{0.956} 
        & \goodres{0.957} 
        & \goodres{0.959} 
        & \goodres{0.956} 
        & \goodres{0.957} 
        & \goodres{0.959} 
        & \goodres{-- 7.4} 
        & \goodres{-- 6.9} 
        \\
        & 20\% 
        & \badres{0.140} 
        & \badres{0.156} 
        & \badres{0.164} 
        & \badres{0.122} 
        & \badres{0.148} 
        & \badres{0.177} 
        \\
        \midrule
        & 5\% 
        & \goodres{0.978} 
        & \goodres{0.979} 
        & \goodres{0.980} 
        & \goodres{0.978} 
        & \goodres{0.979} 
        & \goodres{0.980} 
        \\
        500 
        & 10\% 
        & \goodres{0.957} 
        & \goodres{0.958} 
        & \goodres{0.959} 
        & \goodres{0.957} 
        & \goodres{0.958} 
        & \goodres{0.959} 
        & \goodres{-- 7.5} 
        & \goodres{-- 7.4} 
        \\
        & 20\% 
        & \badres{0.131} 
        & \badres{0.135} 
        & \badres{0.138} 
        & \badres{0.132} 
        & \badres{0.140} 
        & \badres{0.144} 
        \\
        \midrule
        & 5\% 
        & \goodres{0.979} 
        & \goodres{0.979} 
        & \goodres{0.979} 
        & \goodres{0.979} 
        & \goodres{0.979} 
        & \goodres{0.979} 
        \\
        1000 
        & 10\% 
        & \goodres{0.957} 
        & \goodres{0.958} 
        & \goodres{0.958} 
        & \goodres{0.957} 
        & \goodres{0.958} 
        & \goodres{0.958} 
        & \goodres{-- 7.5} 
        & \goodres{-- 7.3} 
        \\
        & 20\% 
        & \badres{0.129} 
        & \badres{0.133} 
        & \badres{0.136} 
        & \badres{0.132} 
        & \badres{0.139} 
        & \badres{0.145} 
        \\
        \bottomrule
    \end{tabular}
\end{table}

\begin{table}[tp]
    \centering
    \caption{\label{tab:DefaultParam}%
    Default values of the model parameters in all numerical experiments. 
    }
    \vspace{0.5em}
    \setlength{\tabcolsep}{3pt}
    \begin{tabular}{llr}
        \toprule
        Symbol & Description & Default Value \\
        \midrule
        $L$  &  number of neurons (Section~\ref{sec:SpikeNetwork}) & $200$ \\
        $K$  &  number of inputs per neuron (Section~\ref{sec:SpikeNetwork}) & $500$ \\
        $\beta$ & input kernel parameter (Section~\ref{sec:Neuron}) & $\taumin$ \\
        $d_\text{min}$  &  minimal axonal delay (Section~\ref{sec:Neuron}) & $0.1 \, \taumin$ \\
        $d_\text{max}$  &  maximal axonal delay (Section~\ref{sec:Neuron}) & $10 \, \taumin$ \\
        $\theta_0$ & nominal threshold (Section~\ref{sec:Threshold}) & $1$  \\
        $\sigma_\theta$ & standard deviation of threshold noise (Section~\ref{sec:Threshold}) & $0.05 \, \theta_0$ \\
        $T$  &  period of firing score (Section~\ref{sec:PrescribedScore} and \ref{app:SpikeTrain}) & $50  \, \taumin$ \\
        $\lambda$ & random-firing rate (\ref{app:SpikeTrain}) & $0.5 / \taumin$ \\
        $\varepsilon_{\mathrm{s}}$ & half-width of firing zone \eqref{eqn:TemplateCondExtraBelow}, \eqref{eqn:TemplateCondSlope} & $0.2 \, \taumin$ \\
        $\zmax$ & maximum potential \eqref{eqn:TemplateCondExtraBelow}  & $0$ \\
        $\dzmin$ & minimal potential slope \eqref{eqn:TemplateCondSlope} & $2 \, \theta_0/\taumin$ \\
        $w_{\mathrm{b}}$ & weight bound \eqref{eqn:WeightsBound} & $0.2 \, \theta_0$ \\
        \bottomrule
    \end{tabular}
\end{table}

The primary result of this paper 
is that exact memorization and stable autonomous reproduction 
is possible (in same range of parameters), as more precisely stated at the beginning of this section.
Some exemplary experimental results proving this are reported 
in \tabref{tab:ThresholdNoiseStabilityNeurons},
which shows the precision (\ref{eqn:Precision}) 
and the recall (\ref{eqn:Recall})
for different values of $L$ (the number of neurons) and $\sigma_\theta$ 
(the standard deviation of the threshold noise).
The values of all other parameters are listed in \tabref{tab:DefaultParam},
which shows the default values of all parameters in all numerical experiments
of this paper.
With $\lambda$ and $T$ as in \tabref{tab:DefaultParam},
the expected total number of spikes (of all neurons, during one period of duration $T$)
is about $13L$ (cf.\ \figref{fig:PMFNumSpikes} in \ref{app:SpikeTrain}).

In these experiments, 
the synaptic weights are computed as described in Section~\ref{sec:compute_weights} (below).
The network is initialized with the exact correct firing times;
then it is left to operate autonomously, with threshold noise as in Section~\ref{sec:Threshold}.
After about five periods, 
the precision 
and the recall 
have reached a steady state, 
which is reported in \tabref{tab:ThresholdNoiseStabilityNeurons} (after 50 periods);
running the simulations longer does not change anything 
(but see Section~\ref{sec:VeryLongTermStability} for extreme time scales).

Every row of \tabref{tab:ThresholdNoiseStabilityNeurons}
summarizes 10 repetitions of the same experiment, 
each with a new random network and a new random prescribed firing score;
each row reports the median, the minimum, and the maximum 
values of (\ref{eqn:Precision}) and (\ref{eqn:Recall})
of all 10 repetitions.

For the two lower levels of threshold noise, 
the precision (\ref{eqn:Precision}) and the recall (\ref{eqn:Recall}) 
always exceed the thresholds~(\ref{eqn:AccuracyThresholdCondition}), 
thus indicating very good reproduction of the prescribed score of spike trains.
For the highest level of threshold noise, the network always fails, which is inevitable 
at some noise level.
(Beyond these essential observations, 
the exact values 
in \tabref{tab:ThresholdNoiseStabilityNeurons}
are of limited interest.)

The last column of \tabref{tab:ThresholdNoiseStabilityNeurons}
shows the minimum and the maximum (among the 10 repetitions)
of the logarithm of the spectral radius $\rhomax$ as in (\ref{eqn:SpecRadCond}),
which does not depend on $\sigma_\theta$.
These values are all negative, showing that condition~(\ref{eqn:SpecRadCond}) is always satisfied.

These experiments 
are representative of large numbers of similar experiments,
with qualitatively similar results.
Some limitations and the role of some parameters will be explored in the remaining sections.

\section{Computing the Synaptic Weights}
\label{sec:compute_weights}

The synaptic weights $w_{1,1},\ldots,w_{L,K}$
are computed to minimize
\begin{equation} \label{eqn:RegL2}
 \sum_{k=1}^K \sum_{\ell=1}^L w_{\ell,k}^2
\end{equation}
subject to the constraints
\begin{IEEEeqnarray}{CCl}
    z_\ell(t) = \theta_0  & \qquad & \text{if $t\in\breve{\set{S}}_\ell$} \label{eqn:TemplateCondCrossing}\\
    z_\ell(t) <  \zmax & & \text{unless $s-\varepsilon_{\mathrm{s}} < t<s+\taumin$ for some $s\in \breve{\set{S}}_\ell$} \label{eqn:TemplateCondExtraBelow}\\
    \dot z_\ell(t) > \dzmin  & & \text{if $s-\varepsilon_{\mathrm{s}} < t<s+\varepsilon_{\mathrm{s}}$ for some $s\in \breve{\set{S}}_\ell$} \label{eqn:TemplateCondSlope}\\
    |w_{\ell, k}| \leq  w_{\mathrm{b}} & & \text{for $k =1,...,K$}   \label{eqn:WeightsBound}
\end{IEEEeqnarray}
for $\ell = 1,\ldots, L,$
where
$x_\ell(t) = \breve x_\ell(t)$ is the prescribed periodic spike train for neuron $\ell$, 
$\breve {\set{S}}_\ell$ is the set of firing times of $\breve x_\ell$,
and $\dot z_\ell(t)$ is the derivative of $z_\ell(t)$ with respect to $t$.
Conditions \eqref{eqn:TemplateCondCrossing}--\eqref{eqn:TemplateCondSlope} 
are illustrated in \figref{fig:Template}.
(The meaningful range of the parameters in \eqref{eqn:TemplateCondExtraBelow}--\eqref{eqn:WeightsBound}
is limited by
$0 < w_{\mathrm{b}} < \theta_0$, $\dzmin > 0$, $\zmax < \theta_0$.)

\begin{figure}[t]
    \centering
        \begin{tikzpicture}
            \tikzset{cross/.style={draw, cross out, minimum size=2*(#1-\pgflinewidth), inner sep=0pt, outer sep=0pt}, cross/.default={1pt}}
            \begin{groupplot}
                [
                    group style={group size=1 by 2, vertical sep=2mm},
                    width=0.9\textwidth,
                    height=0.3\textwidth,
                    tick pos=left,
                    xmin=0, xmax=5.3,
                    ymin=-1, ymax=2,
                    clip=false,
                    xtick={0,1,2,3,4,5,6},
                    xticklabels=\empty,
                    ytick=\empty,
                    ylabel style={at={(0,1)}, anchor=north west, rotate=-90},
                ]   
                \nextgroupplot[ylabel=$z_\ell(t)$]
                \path[name path=topborder] (\pgfkeysvalueof{/pgfplots/xmin},\pgfkeysvalueof{/pgfplots/ymax}) -- (\pgfkeysvalueof{/pgfplots/xmax},\pgfkeysvalueof{/pgfplots/ymax});
                \node[label={[gray] left:$\theta_0$}] at (\pgfkeysvalueof{/pgfplots/xmin}, 1) {};
                \node[label={[gray] left:$\zmax$}] at (\pgfkeysvalueof{/pgfplots/xmin}, 0) {};
                \addplot[gray, dashed, forget plot] coordinates {(\pgfkeysvalueof{/pgfplots/xmin}, 1) (\pgfkeysvalueof{/pgfplots/xmax}, 1)};
                \addplot[gray, dashed, forget plot] coordinates {(\pgfkeysvalueof{/pgfplots/xmin}, 0) (\pgfkeysvalueof{/pgfplots/xmax}, 0)};
                \addplot[blue!60!black, thick, forget plot] coordinates {(1, \pgfkeysvalueof{/pgfplots/ymin}) (1, \pgfkeysvalueof{/pgfplots/ymax})};
                \addplot[blue!60!black, forget plot] coordinates {(0.8, \pgfkeysvalueof{/pgfplots/ymin}) (0.8, \pgfkeysvalueof{/pgfplots/ymax})};
                \addplot[blue!60!black, forget plot] coordinates {(1.2, \pgfkeysvalueof{/pgfplots/ymin}) (1.2, \pgfkeysvalueof{/pgfplots/ymax})};
                \addplot[blue!60!black, thick, forget plot] coordinates {(4.3, \pgfkeysvalueof{/pgfplots/ymin}) (4.3, \pgfkeysvalueof{/pgfplots/ymax})};
                \addplot[blue!60!black, forget plot] coordinates {(4.1, \pgfkeysvalueof{/pgfplots/ymin}) (4.1, \pgfkeysvalueof{/pgfplots/ymax})};
                \addplot[blue!60!black, forget plot] coordinates {(4.5, \pgfkeysvalueof{/pgfplots/ymin}) (4.5, \pgfkeysvalueof{/pgfplots/ymax})};
                \addplot[red!60!black, very thick, samples=100, domain=0:0.8, name path=b1] {exp(-12-x)};
                \addplot[red!60!black, very thick, samples=100, domain=2:4.1, name path=b2, forget plot] {0};
                \addplot[red!10, forget plot] fill between[of=topborder and b1, soft clip={domain=0:0.8}];
                \addplot[red!10, forget plot] fill between[of=topborder and b2, soft clip={domain=2:4.1}];
                \node[cross, green!60!black, very thick, minimum size=5pt] at (1, 1) {};
                \node[cross, green!60!black, very thick, minimum size=5pt] at (4.3, 1) {};
                \draw[<->, blue!60!black] (0.8, \pgfkeysvalueof{/pgfplots/ymax}+0.2) -- node[above, blue!60!black] {$2 \varepsilon_{\mathrm{s}}$} (1.2, \pgfkeysvalueof{/pgfplots/ymax}+0.2);
                \draw[<->, blue!60!black] (4.1, \pgfkeysvalueof{/pgfplots/ymax}+0.2) -- node[above, blue!60!black] {$2 \varepsilon_{\mathrm{s}}$} (4.5, \pgfkeysvalueof{/pgfplots/ymax}+0.2);

                \nextgroupplot[ylabel=$\dot{z}_\ell(t)$]
                \path[name path=bottomborder] (\pgfkeysvalueof{/pgfplots/xmin},\pgfkeysvalueof{/pgfplots/ymin}) -- (\pgfkeysvalueof{/pgfplots/xmax},\pgfkeysvalueof{/pgfplots/ymin});
                \draw[gray, dashed] (\pgfkeysvalueof{/pgfplots/xmin},1) -- (\pgfkeysvalueof{/pgfplots/xmax},1);
                \node[label={[gray] left:$\dzmin$}] at (\pgfkeysvalueof{/pgfplots/xmin}, 1) {};        
                \addplot[blue!60!black, thick, forget plot] coordinates {(1, \pgfkeysvalueof{/pgfplots/ymin}) (1, \pgfkeysvalueof{/pgfplots/ymax})};
                \addplot[blue!60, forget plot] coordinates {(0.8, \pgfkeysvalueof{/pgfplots/ymin}) (0.8, \pgfkeysvalueof{/pgfplots/ymax})};
                \addplot[blue!60, forget plot] coordinates {(1.2, \pgfkeysvalueof{/pgfplots/ymin}) (1.2, \pgfkeysvalueof{/pgfplots/ymax})};
                \addplot[blue!60!black, thick, forget plot] coordinates {(4.3, \pgfkeysvalueof{/pgfplots/ymin}) (4.3, \pgfkeysvalueof{/pgfplots/ymax})};
                \addplot[blue!60, forget plot] coordinates {(4.1, \pgfkeysvalueof{/pgfplots/ymin}) (4.1, \pgfkeysvalueof{/pgfplots/ymax})};
                \addplot[blue!60, forget plot] coordinates {(4.5, \pgfkeysvalueof{/pgfplots/ymin}) (4.5, \pgfkeysvalueof{/pgfplots/ymax})};
                \node[blue!60!black, label={[blue!60!black] below:$s_n$}] at (1, \pgfkeysvalueof{/pgfplots/ymin}) {};
                \node[blue!60!black, label={[blue!60!black] below:$s_{n+1}$}] at (4.3, \pgfkeysvalueof{/pgfplots/ymin}) {};
                \draw[red!60!black, very thick, name path=a1] (0.8, 1.0) -- (1.2, 1.0);
                \draw[red!60!black, very thick, name path=a2] (4.1, 1.0) -- (4.5, 1.0);
                \addplot[red!10, forget plot] fill between[of=bottomborder and a1, soft clip={domain=0.8:1.2}];
                \addplot[red!10, forget plot] fill between[of=bottomborder and a2, soft clip={domain=4.1:4.5}];
            \end{groupplot}
        \end{tikzpicture}  
    \caption{\label{fig:Template}%
    Conditions (\ref{eqn:TemplateCondCrossing})--(\ref{eqn:TemplateCondExtraBelow}) (top) 
    and condition (\ref{eqn:TemplateCondSlope}) (bottom) for two consecutive firing times $s_n$ and $s_{n+1}$. 
    The forbidden regions are shaded in red. 
    The point of firing is marked by a cross. 
    The distance between the tickmarks is $\taumin$.
    }
\end{figure}

Since all firings are prescribed, 
both $z_\ell(t)$ and $\dot z_\ell(t)$ are linear functions of $w_{\ell,1}, \ldots,$ $w_{\ell,K}$.
Therefore, 
minimizing (\ref{eqn:RegL2}) subject to 
\eqref{eqn:TemplateCondCrossing}--\eqref{eqn:WeightsBound}
is a convex linearly constrained optimization problem,
with no coupling between the neurons;
in particular, minimizing \eqref{eqn:RegL2}
decouples into minimizing
\begin{equation} \label{eqn:DecoupledRegL2}
 \sum_{k=1}^K w_{\ell,k}^2
\end{equation}
for each $\ell$.
Note also that \eqref{eqn:TemplateCondCrossing}--\eqref{eqn:WeightsBound} may be infeasible.

One way%
\sfootnote{This was used in an earlier version 
(\href{https://arxiv.org/abs/2408.01166v2}{arXiv:2408.01166v2})
of this paper.} 
to numerically compute such weights (if they exist) is to discretize the continuous time
in (\ref{eqn:TemplateCondCrossing})--(\ref{eqn:TemplateCondSlope}) 
with sufficient temporal resolution and to use standard software for the optimization.
However, in this paper, 
we use a different method 
(described in \ref{app:Optimization})
that turns out to be much faster.

Similar conditions for the potential $z_\ell(t)$ have been used by \cite{LeeAl17},
which, however, did not address temporal stability.
The minimal-slope condition (\ref{eqn:TemplateCondSlope})
appears to be a new ingredient,
which is essential for temporal stability.
Its effect is demonstrated in \tabref{tab:TemplateCondSlope}.
In particular, the first row of \tabref{tab:TemplateCondSlope}
shows that dropping the minimal-slope condition 
makes it impossible for the network to perform properly.
Note that, in this first row, the condition (\ref{eqn:SpecRadCond})
is violated, which means that the memorized content cannot be
stably reproduced even without threshold noise.

\begin{table}[p]
    \centering
    \caption{%
    Effect of the minimum-slope condition (\ref{eqn:TemplateCondSlope}),
    with other parameters as in Table~\ref{tab:DefaultParam}.
    Every row summarizes 10 repetitions of the same experiment.
    Bold type indicates success, faint gray indicates failure.}
    \vspace{0.5em}
    \setlength{\tabcolsep}{3pt}
    \begin{tabular}{c@{\hspace{1.2em}}c@{\hspace{1em}}ccc@{\hspace{1em}}ccc@{\hspace{1.5em}}cc}
        \toprule
        & 
        & \multicolumn{3}{c}{$\pr(50T)$} 
        & \multicolumn{3}{c}{$\rc(50T)$}
        & \multicolumn{2}{c}{$\ln \rhomax$} \\
        $\dzmin \taumin / \theta_0$ 
        & $\sigma_{\theta} / \theta_0$ 
        & $\min$ 
        & $\mathrm{med}$ 
        & $\max$ 
        & $\min$ 
        & $\mathrm{med}$ 
        & $\max$ 
        & $\min$ 
        & $\max$ 
        \\
        \midrule
        & 2\% 
        & \badres{0.170} 
        & \badres{0.197} 
        & \badres{0.280} 
        & \badres{0.116} 
        & \badres{0.134} 
        & \badres{0.144} 
        \\
        0.0 
        & 5\% 
        & \badres{0.166} 
        & \badres{0.213} 
        & \badres{0.276} 
        & \badres{0.124} 
        & \badres{0.135} 
        & \badres{0.148} 
        & \badres{9.0} 
        & \badres{28.9} 
        \\
        & 10\% 
        & \badres{0.189} 
        & \badres{0.219} 
        & \badres{0.255} 
        & \badres{0.108} 
        & \badres{0.128} 
        & \badres{0.145} 
        \\
        \midrule
        & 2\% 
        & \goodres{0.988} 
        & \goodres{0.989} 
        & \goodres{0.989} 
        & \goodres{0.988} 
        & \goodres{0.989} 
        & \goodres{0.989} 
        \\
        0.5 
        & 5\% 
        & \goodres{0.966} 
        & \goodres{0.973} 
        & \goodres{0.974} 
        & \goodres{0.971} 
        & \goodres{0.973} 
        & \goodres{0.974} 
        & \goodres{-- 5.8} 
        & \goodres{-- 4.2}
        \\
        & 10\% 
        & \badres{0.180} 
        & \badres{0.211} 
        & \badres{0.278} 
        & \badres{0.129} 
        & \badres{0.139} 
        & \badres{0.148} 
        \\
        \midrule
        & 2\% 
        & \goodres{0.990} 
        & \goodres{0.990} 
        & \goodres{0.991} 
        & \goodres{0.990} 
        & \goodres{0.990} 
        & \goodres{0.991} 
        \\
        1.0 
        & 5\% 
        & \goodres{0.975} 
        & \goodres{0.976} 
        & \goodres{0.977} 
        & \goodres{0.975} 
        & \goodres{0.976} 
        & \goodres{0.977} 
        & \goodres{-- 6.8} 
        & \goodres{-- 6.2} 
        \\
        & 10\% 
        & \badres{0.204} 
        & \goodres{0.950} 
        & \goodres{0.953} 
        & \badres{0.136} 
        & \goodres{0.950} 
        & \goodres{0.953} 
        \\
        \midrule
        & 2\% 
        & \goodres{0.991} 
        & \goodres{0.991} 
        & \goodres{0.992} 
        & \goodres{0.991} 
        & \goodres{0.991} 
        & \goodres{0.992} 
        \\
        2.0 
        & 5\% 
        & \goodres{0.978} 
        & \goodres{0.979} 
        & \goodres{0.980} 
        & \goodres{0.978} 
        & \goodres{0.979} 
        & \goodres{0.980} 
        & \goodres{-- 7.5} 
        & \goodres{-- 7.0} 
        \\
        & 10\% 
        & \goodres{0.956} 
        & \goodres{0.958} 
        & \goodres{0.959} 
        & \goodres{0.956} 
        & \goodres{0.958} 
        & \goodres{0.959} 
        \\
        \bottomrule
    \end{tabular}
    \label{tab:TemplateCondSlope}
\end{table}

\begin{table}[p]
    \centering
    \caption{%
    Effect of the maximum-level condition (\ref{eqn:TemplateCondExtraBelow}).
    Every row summarizes 10 repetitions of the same experiment.
    Bold type indicates success, faint gray indicates failure.
    The first row coindices with the last row of \tabref{tab:TemplateCondSlope}.}
    \vspace{0.5em}
    \setlength{\tabcolsep}{3pt}
    \begin{tabular}{c@{\hspace{1.2em}}c@{\hspace{1em}}ccc@{\hspace{1em}}ccc@{\hspace{1.5em}}cc}
        \toprule
        & 
        & \multicolumn{3}{c}{$\pr(50T)$} 
        & \multicolumn{3}{c}{$\rc(50T)$}
        & \multicolumn{2}{c}{$\ln \rhomax$} \\
        $\zmax / \theta_0$ 
        & $\sigma_{\theta} / \theta_0$ 
        & $\min$ 
        & $\mathrm{med}$ 
        & $\max$ 
        & $\min$ 
        & $\mathrm{med}$ 
        & $\max$ 
        & $\min$ 
        & $\max$ 
        \\
        \midrule
        & 2\%
        & \goodres{0.991} 
        & \goodres{0.991} 
        & \goodres{0.992} 
        & \goodres{0.991} 
        & \goodres{0.991} 
        & \goodres{0.992} 
        \\
        0 
        & 5\% 
        & \goodres{0.978} 
        & \goodres{0.979} 
        & \goodres{0.980} 
        & \goodres{0.978} 
        & \goodres{0.979} 
        & \goodres{0.980} 
        & \goodres{-- 7.5} 
        & \goodres{-- 7.0} 
        \\
        & 10\% 
        & \goodres{0.956} 
        & \goodres{0.958} 
        & \goodres{0.959} 
        & \goodres{0.956} 
        & \goodres{0.958} 
        & \goodres{0.959} 
        \\
        \midrule
        & 2\% 
        & \goodres{0.985} 
        & \goodres{0.986} 
        & \goodres{0.986} 
        & \goodres{0.985} 
        & \goodres{0.986} 
        & \goodres{0.986} 
        \\
        0.5 
        & 5\% 
        & \goodres{0.964} 
        & \goodres{0.966} 
        & \goodres{0.967} 
        & \goodres{0.964} 
        & \goodres{0.966} 
        & \goodres{0.967} 
        & \goodres{-- 7.2} 
        & \goodres{-- 6.9} 
        \\
        & 10\% 
        & \badres{0.152} 
        & \goodres{0.929} 
        & \goodres{0.932} 
        & \badres{0.204} 
        & \goodres{0.930} 
        & \goodres{0.932} 
        \\
        \midrule
        & 2\% 
        & \goodres{0.984} 
        & \goodres{0.985} 
        & \goodres{0.985} 
        & \goodres{0.984} 
        & \goodres{0.985} 
        & \goodres{0.985} 
        \\
        0.8 
        & 5\% 
        & \badres{0.137} 
        & \badres{0.144} 
        & \badres{0.168} 
        & \badres{0.244} 
        & \badres{0.292} 
        & \badres{0.338} 
        & \goodres{-- 7.5} 
        & \goodres{-- 7.1} 
        \\
        & 10\% 
        & \badres{0.140} 
        & \badres{0.156} 
        & \badres{0.194} 
        & \badres{0.237} 
        & \badres{0.277} 
        & \badres{0.324} 
        \\
        \bottomrule
    \end{tabular}
    \label{tab:TemplateCondExtraBelow}
\end{table}

\tabref{tab:TemplateCondExtraBelow} shows the effect of the condition (\ref{eqn:TemplateCondExtraBelow}).
Note that the minimum-slope condition (\ref{eqn:TemplateCondSlope})
is active (cf.\ Table~\ref{tab:DefaultParam})
and guarantees (\ref{eqn:SpecRadCond}).
However, 
increasing $\zmax$ decreases the robustness 
of the network against threshold noise.
On the other hand, choosing $\zmax$ too low may result in 
(\ref{eqn:TemplateCondCrossing})--(\ref{eqn:WeightsBound})
to be infeasible (but this is not documented in \tabref{tab:TemplateCondExtraBelow}).%

From the experiments in 
\tabsref{tab:TemplateCondSlope} and~\ref{tab:TemplateCondExtraBelow}, 
and from many other experiments,
we observe that satisfying \eqref{eqn:TemplateCondCrossing}--\eqref{eqn:WeightsBound}
(with sufficiently large $\dzmin$ and sufficiently small $w_{\mathrm{b}}$ and $\zmax$,
for all neurons simultaneously)
suffices to guarantee temporal stability
at some (positive) level of threshold noise.
The feasibility of (\ref{eqn:TemplateCondCrossing})--(\ref{eqn:WeightsBound})
can therefore be used as a (conservative) proxy 
for stable memorization.

\begin{figure}
\centering
\begin{tikzpicture}
    \begin{axis}[
        width=0.9\textwidth,
        height=0.4\textwidth,
        tick pos=left,
        xlabel={$T / \taumin$},
        ylabel={probability of feasibility},
        grid={major},
        xmin=55, xmax=165,
        ymin=0, ymax=1,
        legend style={at={(0.5, 1.05)}, anchor=south, /tikz/every even column/.style={column sep=8mm}},
        legend columns=3,
      ]
  
      \addlegendimage{red!80!black, thick}
      \addlegendentry{$K=500$}
      \addlegendimage{green!60!black, thick}
      \addlegendentry{$K=600$}
      \addlegendimage{blue!80!black, thick}
      \addlegendentry{$K=700$}
  
      \addplot[red!80!black, domain=40:200, samples=200, thick, smooth] plot (\x, {1/(1 + exp(-22.87419747 + 0.30707937 * \x))}); 
      \addplot[only marks, red!80!black, mark=*, mark size=1.5pt] coordinates {(55, 1.0) (60, 1.0) (65, 0.9) (70, 0.8) (75, 0.5) (80, 0.2) (85, 0.0) (90, 0.0)};
      
      \addplot[green!60!black, domain=40:200, samples=200, thick, smooth] plot (\x, {1/(1 + exp(-26.67636268 + 0.25398932 * \x))}); 
      \addplot[only marks, green!60!black, mark=*, mark size=1.5pt] coordinates {(85, 1.0) (90, 1.0) (95, 0.9) (100, 0.6) (105, 0.7) (110, 0.3) (115, 0.0) (120, 0.0) (125, 0.0)};
  
      \addplot[blue!80!black, domain=40:200, samples=200, thick, smooth] plot (\x, {1/(1 + exp(-34.21909744 + 0.2486457 * \x))}); 
      \addplot[only marks, blue!80!black, mark=*, mark size=1.5pt] coordinates {(120, 1.0) (125, 1.0) (130, 0.8) (135, 0.6) (140, 0.4) (145, 0.2) (150, 0.0) (155, 0.0) (160, 0.0)};
    \end{axis}
\end{tikzpicture}
\caption{\label{fig:CapacityData}%
Empirical probability of feasibility vs.\ the period~$T$, 
averaged over 10 repetitions of the same experiment
(each with a new random network and a new prescribed random spike train,
with default parameters as in Table~\ref{tab:DefaultParam}). 
The solid lines are logistic regressions to the simulation results.
Each neuron fires about 25 times (on average) per 100$\,\taumin$;
for $T=100\,\taumin$, there are about 5000 firings (per period) in total.
}
\vspace{\floatsep}

\begin{tikzpicture}
    \begin{axis}[
        width=0.98\textwidth,
        height=0.4\textwidth,
        tick pos=left,
        xlabel={$T / \taumin$},
        ylabel={probability of feasibility},
        xmajorgrids,
        ymajorgrids,
        xmin=50, xmax=200,
        ymin=0, ymax=1,
        legend style={at={(0.5, 1.05)}, anchor=south, /tikz/every even column/.style={column sep=6mm}},
        legend columns=2,
      ]
  
      \addlegendimage{red!80!black, thick}
      \addlegendentry{$K=500$}
      \addlegendimage{thick, smooth, dashed}
      \addlegendentry{$\lambda \taumin = 0.2$ ($\simeq$ 15 firings per neuron and per $100 \, \taumin$)}
      \addlegendimage{green!60!black, thick}
      \addlegendentry{$K=600$}
      \addlegendimage{thick, smooth}
      \addlegendentry{$\lambda \taumin = 0.5$ ($\simeq$ 25 firings per neuron and per $100 \, \taumin$)}
      \addlegendimage{blue!80!black, thick}
      \addlegendentry{$K=700$}
      \addlegendimage{thick, smooth, dotted}
      \addlegendentry{$\lambda \taumin=1.0$ ($\simeq$ 35 firings per neuron and per $100 \, \taumin$)}
  
      \addplot[red!80!black, domain=40:200, samples=200, thick, smooth, dashed] plot (\x, {1/(1 + exp(-9.01667792 + 0.1126507 * \x))}); 
      \addplot[red!80!black, domain=40:200, samples=200, thick, smooth] plot (\x, {1/(1 + exp(-22.87419747 + 0.30707937 * \x))}); 
      \addplot[red!80!black, domain=40:200, samples=200, thick, smooth, dotted] plot (\x, {1/(1 + exp(-42.60000761 + 0.60465765 * \x))}); 
  
      \addplot[green!60!black, domain=40:200, samples=200, thick, smooth, dashed] plot (\x, {1/(1 + exp(-17.34377968 + 0.14731281 * \x))}); 
      \addplot[green!60!black, domain=40:200, samples=200, thick, smooth] plot (\x, {1/(1 + exp(-26.67636268 + 0.25398932 * \x))}); 
      \addplot[green!60!black, domain=40:200, samples=200, thick, smooth, dotted] plot (\x, {1/(1 + exp(-38.19563635 + 0.40434648 * \x))}); 
  
      \addplot[blue!80!black, domain=40:200, samples=200, thick, smooth, dashed] plot (\x, {1/(1 + exp(-22.14287497 + 0.13407951 * \x))}); 
      \addplot[blue!80!black, domain=40:200, samples=200, thick, smooth] plot (\x, {1/(1 + exp(-34.21909744 + 0.2486457 * \x))}); 
      \addplot[blue!80!black, domain=40:200, samples=200, thick, smooth, dotted] plot (\x, {1/(1 + exp(-38.26659948 + 0.31027932 * \x))}); 
    \end{axis}
    \end{tikzpicture}
\caption{\label{fig:CapacityForK}%
Empirical probability of feasibility vs.\ the period~$T$,
for different values of the Poisson parameter~$\lambda$,
with other details as in Fig.~\ref{fig:CapacityData}.
Each line is a logistic regression to the simulation results 
(not shown here).}
\end{figure}

\section{Memory Capacity}
\label{sec:Capacity}

Clearly, for any fixed network,
there must be an upper limit on the maximal period $T$ of the memorizable content.
Some pertinent (and typical) experimental results are reported in 
\figsref{fig:CapacityData} and \ref{fig:CapacityForK}. 
In these experiments, the feasibility of (\ref{eqn:TemplateCondCrossing})--(\ref{eqn:WeightsBound})
is used as a proxy for stable memorization (as explained above). 
\figsref{fig:CapacityData} and \ref{fig:CapacityForK} 
both show the probability
of this feasibility as a function of $T$ for different numbers of inputs per neuron $K$.
\figref{fig:CapacityForK} also shows the influence of 
the Poisson parameter $\lambda$, which determines the average number 
of spikes per neuron 
(cf.~\figref{fig:PMFNumSpikes} in \ref{app:SpikeTrain}).

From many more such experiments (not shown here), 
we conclude that, for fixed~$\lambda$, $K$ is indeed the primary pertinent parameter. 
Both $L$ (= the number of neurons) 
and $\dmax - \dmin$ (= the axonal delay spread)
just need to be sufficiently large (and can be traded against each other).

Note that \figref{fig:CapacityForK} 
suggests that the maximal length $T$ of stably memorizable content grows at least linearly with $K$.
(A similar scaling%
\sfootnote{%
This scaling is also consistent with the theoretical result of \cite{Motzkin55} 
concerning the probability of solvability of systems of linear inequalities, 
applied to (\ref{eqn:TemplateCondCrossing})--(\ref{eqn:TemplateCondSlope})
with sufficiently fine temporal discretization.
} 
was mathematically proved in a discrete-time setting by \cite{MurerLoeliger20}.)

The numbers reported in \figsref{fig:CapacityData} and \ref{fig:CapacityForK}
depend, or course, on the values of the template parameters,
which were not optimized for this purpose. 
Therefore, \figsref{fig:CapacityData} and \ref{fig:CapacityForK} 
are conservative estimates of the capacity.

Finally, we note that partitioning the prescribed score of spike trains 
into several periodic scores with periods $T_1,\ldots,T_n \gg \taumin$
of total duration $T = T_1 + \ldots + T_n$
does not affect the feasibility of (\ref{eqn:TemplateCondCrossing})--(\ref{eqn:WeightsBound})
and, consequently, the total capacity of the network.
Simulation results 
with such a partitioning 
are given in Section~\ref{sec:TwoDisjointMemories}.

\section{Associative Recall} 
\label{sec:AssociativeRecall}

As expected, the memorized spike score(s) can be recalled by noisy partial prompts.
Some pertinent experiments are described below.
In these experiments,
a fraction $\alpha$ of the neurons does not (or not always) operate in its normal mode:
these $\alpha L$ neurons 
are forced to ignore their input and to produce a noisy version 
of their (prescribed and) memorized spike train instead.
These noisy excitations are created by shifting each spike of the prescribed spike train 
by random zero-mean Gaussian jitter with variance $\sigma_{\mathrm{s}}^2$ 
while preserving the refractory period $\taumin$ between spikes,
as detailed in \ref{appsec:PromptJitter}.

\subsection{Recall from Quiet State}
\label{sec:Prompts}

In this experiment, the network is prepared as described in 
Section~\ref{sec:compute_weights}, with parameters as in \tabref{tab:DefaultParam}.
The network is started in a state of rest 
(with no firings and all neuron potentials set to zero), 
but with a fraction $\alpha$ of neurons 
forced to provide 
a noisy version of the memorized content as described above.

Numerical results of five repetitions 
(each with a different network and spike score) of such an experiment 
with $\alpha = 55\%$ are shown in \figref{fig:quiet}. 
Note that the unforced neurons eventually lock in to the memorized content, 
with higher temporal accuracy than the (noisy) excitation.

\begin{figure}[t]
    \centering
    \begin{tikzpicture}
        \begin{axis}[
            xlabel=$t$,
            ylabel=$\min{\{\mathrm{pr}(t), \mathrm{re}(t)\}}$,
            width=0.9\textwidth,
            height=0.4\textwidth,
            tick pos=left,
            xmin=0, xmax=8,
            ymin=0.5, ymax=1,
            xtick distance=1,
            xticklabels=\empty,
            ytick distance=0.25,
            grid=both,
          ]

        \addplot[blue!60!black, thick, mark=*] plot coordinates {(0, 0.633145139) (1, 0.754823222) (2, 0.946667302) (3, 0.970754243) (4, 0.970742314) (5, 0.97074232) (6, 0.97074232) (7, 0.97074232) (8, 0.97074232) (9, 0.97074232) (10, 0.97074232) (11, 0.97074232) (12, 0.97074232) (13, 0.97074232) (14, 0.97074232) (15, 0.97074232) (16, 0.97074232) (17, 0.97074232) (18, 0.97074232) (19, 0.97074232) (20, 0.97074232) (21, 0.97074232) (22, 0.97074232) (23, 0.97074232) (24, 0.97074232) (25, 0.97074232) (26, 0.97074232) (27, 0.97074232) (28, 0.97074232) (29, 0.97074232) (30, 0.97074232) (31, 0.97074232) (32, 0.97074232) (33, 0.97074232) (34, 0.97074232) (35, 0.97074232) (36, 0.97074232) (37, 0.97074232) (38, 0.97074232) (39, 0.97074232) (40, 0.97074232) (41, 0.97074232) (42, 0.97074232) (43, 0.97074232) (44, 0.97074232) (45, 0.97074232) (46, 0.97074232) (47, 0.97074232) (48, 0.97074232) (49, 0.97074232) (50, 0.97074232) (51, 0.97074232) (52, 0.97074232) (53, 0.97074232) (54, 0.97074232) (55, 0.97074232) (56, 0.97074232) (57, 0.97074232) (58, 0.97074232) (59, 0.97074232) (60, 0.97074232) (61, 0.97074232) (62, 0.97074232) (63, 0.97074232) (64, 0.97074232) (65, 0.97074232) (66, 0.97074232) (67, 0.97074232) (68, 0.97074232) (69, 0.97074232) (70, 0.97074232) (71, 0.97074232) (72, 0.97074232) (73, 0.97074232) (74, 0.97074232) (75, 0.97074232) (76, 0.97074232) (77, 0.97074232) (78, 0.97074232) (79, 0.97074232) (80, 0.97074232) (81, 0.97074232) (82, 0.97074232) (83, 0.97074232) (84, 0.97074232) (85, 0.97074232) (86, 0.97074232) (87, 0.97074232) (88, 0.97074232) (89, 0.97074232) (90, 0.97074232) (91, 0.97074232) (92, 0.97074232) (93, 0.97074232) (94, 0.97074232) (95, 0.97074232) (96, 0.97074232) (97, 0.97074232) (98, 0.97074232) (99, 0.97074232)};
        \addplot[blue!60!black, thick, mark=*] plot coordinates {(0, 0.642637318) (1, 0.659628026) (2, 0.718383162) (3, 0.80394608) (4, 0.961987647) (5, 0.968948933) (6, 0.968952095) (7, 0.968952099) (8, 0.968952099) (9, 0.968952099) (10, 0.968952099) (11, 0.968952099) (12, 0.968952099) (13, 0.968952099) (14, 0.968952099) (15, 0.968952099) (16, 0.968952099) (17, 0.968952099) (18, 0.968952099) (19, 0.968952099) (20, 0.968952099) (21, 0.968952099) (22, 0.968952099) (23, 0.968952099) (24, 0.968952099) (25, 0.968952099) (26, 0.968952099) (27, 0.968952099) (28, 0.968952099) (29, 0.968952099) (30, 0.968952099) (31, 0.968952099) (32, 0.968952099) (33, 0.968952099) (34, 0.968952099) (35, 0.968952099) (36, 0.968952099) (37, 0.968952099) (38, 0.968952099) (39, 0.968952099) (40, 0.968952099) (41, 0.968952099) (42, 0.968952099) (43, 0.968952099) (44, 0.968952099) (45, 0.968952099) (46, 0.968952099) (47, 0.968952099) (48, 0.968952099) (49, 0.968952099) (50, 0.968952099) (51, 0.968952099) (52, 0.968952099) (53, 0.968952099) (54, 0.968952099) (55, 0.968952099) (56, 0.968952099) (57, 0.968952099) (58, 0.968952099) (59, 0.968952099) (60, 0.968952099) (61, 0.968952099) (62, 0.968952099) (63, 0.968952099) (64, 0.968952099) (65, 0.968952099) (66, 0.968952099) (67, 0.968952099) (68, 0.968952099) (69, 0.968952099) (70, 0.968952099) (71, 0.968952099) (72, 0.968952099) (73, 0.968952099) (74, 0.968952099) (75, 0.968952099) (76, 0.968952099) (77, 0.968952099) (78, 0.968952099) (79, 0.968952099) (80, 0.968952099) (81, 0.968952099) (82, 0.968952099) (83, 0.968952099) (84, 0.968952099) (85, 0.968952099) (86, 0.968952099) (87, 0.968952099) (88, 0.968952099) (89, 0.968952099) (90, 0.968952099) (91, 0.968952099) (92, 0.968952099) (93, 0.968952099) (94, 0.968952099) (95, 0.968952099) (96, 0.968952099) (97, 0.968952099) (98, 0.968952099) (99, 0.968952099)};
        \addplot[blue!60!black, thick, mark=*] plot coordinates {(0, 0.567597742) (1, 0.728009467) (2, 0.930065591) (3, 0.971316175) (4, 0.971178052) (5, 0.971332932) (6, 0.971178063) (7, 0.971332932) (8, 0.971178063) (9, 0.971332932) (10, 0.971178063) (11, 0.971332932) (12, 0.971178063) (13, 0.971332932) (14, 0.971178063) (15, 0.971332932) (16, 0.971178063) (17, 0.971332932) (18, 0.971178063) (19, 0.971332932) (20, 0.971178063) (21, 0.971332932) (22, 0.971178063) (23, 0.971332932) (24, 0.971178063) (25, 0.971332932) (26, 0.971178063) (27, 0.971332932) (28, 0.971178063) (29, 0.971332932) (30, 0.971178063) (31, 0.971332932) (32, 0.971178063) (33, 0.971332932) (34, 0.971178063) (35, 0.971332932) (36, 0.971178063) (37, 0.971332932) (38, 0.971178063) (39, 0.971332932) (40, 0.971178063) (41, 0.971332932) (42, 0.971178063) (43, 0.971332932) (44, 0.971178063) (45, 0.971332932) (46, 0.971178063) (47, 0.971332932) (48, 0.971178063) (49, 0.971332932) (50, 0.971178063) (51, 0.971332932) (52, 0.971178063) (53, 0.971332932) (54, 0.971178063) (55, 0.971332932) (56, 0.971178063) (57, 0.971332932) (58, 0.971178063) (59, 0.971332932) (60, 0.971178063) (61, 0.971332932) (62, 0.971178063) (63, 0.971332932) (64, 0.971178063) (65, 0.971332932) (66, 0.971178063) (67, 0.971332932) (68, 0.971178063) (69, 0.971332932) (70, 0.971178063) (71, 0.971332932) (72, 0.971178063) (73, 0.971332932) (74, 0.971178063) (75, 0.971332932) (76, 0.971178063) (77, 0.971332932) (78, 0.971178063) (79, 0.971332932) (80, 0.971178063) (81, 0.971332932) (82, 0.971178063) (83, 0.971332932) (84, 0.971178063) (85, 0.971332932) (86, 0.971178063) (87, 0.971332932) (88, 0.971178063) (89, 0.971332932) (90, 0.971178063) (91, 0.971332932) (92, 0.971178063) (93, 0.971332932) (94, 0.971178063) (95, 0.971332932) (96, 0.971178063) (97, 0.971332932) (98, 0.971178063) (99, 0.971332932)};
        \addplot[blue!60!black, thick, mark=*] plot coordinates {(0, 0.585583118) (1, 0.665462033) (2, 0.764959084) (3, 0.948691988) (4, 0.972514763) (5, 0.972518107) (6, 0.972518106) (7, 0.972518106) (8, 0.972518106) (9, 0.972518106) (10, 0.972518106) (11, 0.972518106) (12, 0.972518106) (13, 0.972518106) (14, 0.972518106) (15, 0.972518106) (16, 0.972518106) (17, 0.972518106) (18, 0.972518106) (19, 0.972518106) (20, 0.972518106) (21, 0.972518106) (22, 0.972518106) (23, 0.972518106) (24, 0.972518106) (25, 0.972518106) (26, 0.972518106) (27, 0.972518106) (28, 0.972518106) (29, 0.972518106) (30, 0.972518106) (31, 0.972518106) (32, 0.972518106) (33, 0.972518106) (34, 0.972518106) (35, 0.972518106) (36, 0.972518106) (37, 0.972518106) (38, 0.972518106) (39, 0.972518106) (40, 0.972518106) (41, 0.972518106) (42, 0.972518106) (43, 0.972518106) (44, 0.972518106) (45, 0.972518106) (46, 0.972518106) (47, 0.972518106) (48, 0.972518106) (49, 0.972518106) (50, 0.972518106) (51, 0.972518106) (52, 0.972518106) (53, 0.972518106) (54, 0.972518106) (55, 0.972518106) (56, 0.972518106) (57, 0.972518106) (58, 0.972518106) (59, 0.972518106) (60, 0.972518106) (61, 0.972518106) (62, 0.972518106) (63, 0.972518106) (64, 0.972518106) (65, 0.972518106) (66, 0.972518106) (67, 0.972518106) (68, 0.972518106) (69, 0.972518106) (70, 0.972518106) (71, 0.972518106) (72, 0.972518106) (73, 0.972518106) (74, 0.972518106) (75, 0.972518106) (76, 0.972518106) (77, 0.972518106) (78, 0.972518106) (79, 0.972518106) (80, 0.972518106) (81, 0.972518106) (82, 0.972518106) (83, 0.972518106) (84, 0.972518106) (85, 0.972518106) (86, 0.972518106) (87, 0.972518106) (88, 0.972518106) (89, 0.972518106) (90, 0.972518106) (91, 0.972518106) (92, 0.972518106) (93, 0.972518106) (94, 0.972518106) (95, 0.972518106) (96, 0.972518106) (97, 0.972518106) (98, 0.972518106) (99, 0.972518106)};
        \addplot[blue!60!black, thick, mark=*] plot coordinates {(0, 0.579382024) (1, 0.634385234) (2, 0.697622417) (3, 0.875174808) (4, 0.975776217) (5, 0.975628213) (6, 0.975628064) (7, 0.975628063) (8, 0.975628063) (9, 0.975628063) (10, 0.975628063) (11, 0.975628063) (12, 0.975628063) (13, 0.975628063) (14, 0.975628063) (15, 0.975628063) (16, 0.975628063) (17, 0.975628063) (18, 0.975628063) (19, 0.975628063) (20, 0.975628063) (21, 0.975628063) (22, 0.975628063) (23, 0.975628063) (24, 0.975628063) (25, 0.975628063) (26, 0.975628063) (27, 0.975628063) (28, 0.975628063) (29, 0.975628063) (30, 0.975628063) (31, 0.975628063) (32, 0.975628063) (33, 0.975628063) (34, 0.975628063) (35, 0.975628063) (36, 0.975628063) (37, 0.975628063) (38, 0.975628063) (39, 0.975628063) (40, 0.975628063) (41, 0.975628063) (42, 0.975628063) (43, 0.975628063) (44, 0.975628063) (45, 0.975628063) (46, 0.975628063) (47, 0.975628063) (48, 0.975628063) (49, 0.975628063) (50, 0.975628063) (51, 0.975628063) (52, 0.975628063) (53, 0.975628063) (54, 0.975628063) (55, 0.975628063) (56, 0.975628063) (57, 0.975628063) (58, 0.975628063) (59, 0.975628063) (60, 0.975628063) (61, 0.975628063) (62, 0.975628063) (63, 0.975628063) (64, 0.975628063) (65, 0.975628063) (66, 0.975628063) (67, 0.975628063) (68, 0.975628063) (69, 0.975628063) (70, 0.975628063) (71, 0.975628063) (72, 0.975628063) (73, 0.975628063) (74, 0.975628063) (75, 0.975628063) (76, 0.975628063) (77, 0.975628063) (78, 0.975628063) (79, 0.975628063) (80, 0.975628063) (81, 0.975628063) (82, 0.975628063) (83, 0.975628063) (84, 0.975628063) (85, 0.975628063) (86, 0.975628063) (87, 0.975628063) (88, 0.975628063) (89, 0.975628063) (90, 0.975628063) (91, 0.975628063) (92, 0.975628063) (93, 0.975628063) (94, 0.975628063) (95, 0.975628063) (96, 0.975628063) (97, 0.975628063) (98, 0.975628063) (99, 0.975628063)};
        \addplot[gray, thick, dashed] plot coordinates {(0, 0.922) (99, 0.922)};
    
        \end{axis}
      \end{tikzpicture}
    \caption{\label{fig:quiet}%
        Recall from quiet state with $\alpha = 55\%$ forced neurons,
        which are forced to produce the memorized firings with 
        jitter parameter $\sigma_{\mathrm{s}} / \taumin = 0.05$.
        The solid lines show the precision or the recall (whichever is smaller) 
        of the autonomous (i.e., unforced) neurons
        for five different simulations, each with a different network and spike score.
        The horizontal dashed line shows the average precision/recall of the forced neurons. 
        The distance between the tickmarks is~$T$.
    }
\end{figure}

\subsection{Switching Between Memorized Content}
\label{sec:TwoDisjointMemories}

In this experiment, the memorized random score of firings 
consists of two segments $\breve{x}_R$ and $\breve{x}_B$, 
each of which is periodic with period $T=50 \, \taumin$
(cf.\ the last paragraph of Section~\ref{sec:Capacity}).
The other parameters are as in Table~\ref{tab:DefaultParam},
except that we here choose $w_{\mathrm{b}} = 0.5 \, \theta_0$.
A fraction $\alpha$ of neurons (henceforth called the forceable neurons) 
is sometimes forced to produce noisy memorized contents as in Section~\ref{sec:Prompts}.

The network is started in a state of rest, 
with no firings and all neuron potentials set to zero.
The simulation proceeds in different phases of duration $T$. 
In each such phase, the network operates in one of the following three modes:
\begin{itemize}
    \item[(1)] Autonomously as in Section~\ref{sec:Network}.
    \item[(2)] The forceable neurons are forced to produce a noisy version of $\breve{x}_R$.
    \item[(3)] The forceable neurons are forced to produce a noisy version of $\breve{x}_B$.
\end{itemize}

\begin{figure}[t]
    \centering
    \begin{tikzpicture}
        \begin{axis}[xlabel=$t$, ylabel=$\min{\{\mathrm{pr}(t), \mathrm{re}(t)\}}$, width=14cm, height=6cm, tick pos=left, xmin=0, xmax=100, ymin=0, ymax=1, xtick distance=10, xticklabels=\empty, ytick distance=0.25, clip=false, grid=both, legend style={at={(0.5,1.2)}, anchor=south, /tikz/every even column/.style={column sep=8pt,}}, legend cell align={left}, legend columns=3]

        \addlegendimage{thick, solid}
        \addlegendentry{all neurons}

        \addlegendimage{thick, dashed}
        \addlegendentry{forceable neurons}

        \addlegendimage{thick, dashdotted}
        \addlegendentry{autonomous neurons}

        \addplot[blue!60!black, thick, opacity=0.6] plot coordinates {(0, 0.810706698) (1, 0.831709217) (2, 0.8368772) (3, 0.830141513) (4, 0.831111018) (5, 0.846611305) (6, 0.86842627) (7, 0.917883507) (8, 0.93171242) (9, 0.931665241) (10, 0.974999353) (11, 0.977569789) (12, 0.977570353) (13, 0.977570368) (14, 0.977570368) (15, 0.977575744) (16, 0.977563577) (17, 0.977561544) (18, 0.977670467) (19, 0.977575094) (20, 0.148290137) (21, 0.148081673) (22, 0.147722023) (23, 0.145526505) (24, 0.145085916) (25, 0.148377953) (26, 0.149319881) (27, 0.146363052) (28, 0.144470006) (29, 0.144330063) (30, 0.144700496) (31, 0.144811265) (32, 0.144793998) (33, 0.144802687) (34, 0.144802708) (35, 0.144740327) (36, 0.144798725) (37, 0.144744755) (38, 0.144800879) (39, 0.146462677) (40, 0.804400258) (41, 0.818335565) (42, 0.839481412) (43, 0.895608834) (44, 0.931700607) (45, 0.931665238) (46, 0.931665239) (47, 0.931665239) (48, 0.931665239) (49, 0.931665239) (50, 0.974999353) (51, 0.977569789) (52, 0.977570353) (53, 0.977570368) (54, 0.977570368) (55, 0.977575744) (56, 0.977563577) (57, 0.977561544) (58, 0.977670467) (59, 0.977575094) (60, 0.148290137) (61, 0.148081673) (62, 0.147722023) (63, 0.145526505) (64, 0.145085916) (65, 0.148377953) (66, 0.149319881) (67, 0.146363052) (68, 0.144470006) (69, 0.144330063) (70, 0.144700496) (71, 0.144811265) (72, 0.144793998) (73, 0.144802687) (74, 0.144802708) (75, 0.144740327) (76, 0.144798725) (77, 0.144744755) (78, 0.144800879) (79, 0.146462677) (80, 0.804400258) (81, 0.818335565) (82, 0.839481412) (83, 0.895608834) (84, 0.931700607) (85, 0.931665238) (86, 0.931665239) (87, 0.931665239) (88, 0.931665239) (89, 0.931665239) (90, 0.974999353) (91, 0.977569789) (92, 0.977570353) (93, 0.977570368) (94, 0.977570368) (95, 0.977575744) (96, 0.977563577) (97, 0.977561544) (98, 0.977670467) (99, 0.977575094)};
        \addplot[red!60!black, thick, opacity=0.6] plot coordinates {(0, 0.143945115) (1, 0.143949129) (2, 0.146340879) (3, 0.147329421) (4, 0.145217464) (5, 0.14324979) (6, 0.146089228) (7, 0.145011057) (8, 0.144103597) (9, 0.144087495) (10, 0.143818411) (11, 0.143978548) (12, 0.143976029) (13, 0.143976044) (14, 0.143976044) (15, 0.143976044) (16, 0.143973294) (17, 0.143977752) (18, 0.144007784) (19, 0.142872475) (20, 0.80646783) (21, 0.816865034) (22, 0.839028306) (23, 0.846089566) (24, 0.853490595) (25, 0.83797858) (26, 0.843601829) (27, 0.89688881) (28, 0.930811995) (29, 0.929689674) (30, 0.973642101) (31, 0.97685923) (32, 0.976898005) (33, 0.976897013) (34, 0.976896981) (35, 0.976737193) (36, 0.976919968) (37, 0.976754034) (38, 0.976929613) (39, 0.976939949) (40, 0.148806633) (41, 0.145138004) (42, 0.14487262) (43, 0.146237599) (44, 0.144110357) (45, 0.144087494) (46, 0.144087495) (47, 0.144087495) (48, 0.144087495) (49, 0.144087495) (50, 0.143818411) (51, 0.143978548) (52, 0.143976029) (53, 0.143976044) (54, 0.143976044) (55, 0.143976044) (56, 0.143973294) (57, 0.143977752) (58, 0.144007784) (59, 0.142872475) (60, 0.80646783) (61, 0.816865034) (62, 0.839028306) (63, 0.846089566) (64, 0.853490595) (65, 0.83797858) (66, 0.843601829) (67, 0.89688881) (68, 0.930811995) (69, 0.929689674) (70, 0.973642101) (71, 0.97685923) (72, 0.976898005) (73, 0.976897013) (74, 0.976896981) (75, 0.976737193) (76, 0.976919968) (77, 0.976754034) (78, 0.976929613) (79, 0.976939949) (80, 0.148806633) (81, 0.145138004) (82, 0.14487262) (83, 0.146237599) (84, 0.144110357) (85, 0.144087494) (86, 0.144087495) (87, 0.144087495) (88, 0.144087495) (89, 0.144087495) (90, 0.143818411) (91, 0.143978548) (92, 0.143976029) (93, 0.143976044) (94, 0.143976044) (95, 0.143976044) (96, 0.143973294) (97, 0.143977752) (98, 0.144007784) (99, 0.14392087)};

        \addplot[blue!60!black, thick, dashdotted] plot coordinates {(0, 0.499861691) (1, 0.565066709) (2, 0.60113756) (3, 0.58440781) (4, 0.570309764) (5, 0.62249196) (6, 0.709960366) (7, 0.908200978) (8, 0.962888579) (9, 0.962705144) (10, 0.974663046) (11, 0.97792349) (12, 0.977930639) (13, 0.977930628) (14, 0.977930628) (15, 0.977930628) (16, 0.97788437) (17, 0.977877459) (18, 0.977842143) (19, 0.977868135) (20, 0.183224668) (21, 0.168947535) (22, 0.169108636) (23, 0.164678001) (24, 0.159491546) (25, 0.163088182) (26, 0.16827178) (27, 0.165310574) (28, 0.165502795) (29, 0.165462976) (30, 0.165464455) (31, 0.165095665) (32, 0.165172886) (33, 0.16514345) (34, 0.165143403) (35, 0.165193938) (36, 0.165142263) (37, 0.165151971) (38, 0.165142914) (39, 0.16515194) (40, 0.453625995) (41, 0.509968799) (42, 0.594008514) (43, 0.821004385) (44, 0.96286385) (45, 0.962705133) (46, 0.962705131) (47, 0.962705131) (48, 0.962705131) (49, 0.962705131) (50, 0.974663046) (51, 0.97792349) (52, 0.977930639) (53, 0.977930628) (54, 0.977930628) (55, 0.977930628) (56, 0.97788437) (57, 0.977877459) (58, 0.977842143) (59, 0.977868135) (60, 0.183224668) (61, 0.168947535) (62, 0.169108636) (63, 0.164678001) (64, 0.159491546) (65, 0.163088182) (66, 0.16827178) (67, 0.165310574) (68, 0.165502795) (69, 0.165462976) (70, 0.165464455) (71, 0.165095665) (72, 0.165172886) (73, 0.16514345) (74, 0.165143403) (75, 0.165193938) (76, 0.165142263) (77, 0.165151971) (78, 0.165142914) (79, 0.16515194) (80, 0.453625995) (81, 0.509968799) (82, 0.594008514) (83, 0.821004385) (84, 0.96286385) (85, 0.962705133) (86, 0.962705131) (87, 0.962705131) (88, 0.962705131) (89, 0.962705131) (90, 0.974663046) (91, 0.97792349) (92, 0.977930639) (93, 0.977930628) (94, 0.977930628) (95, 0.977930628) (96, 0.97788437) (97, 0.977877459) (98, 0.977842143) (99, 0.977868135)};
        \addplot[red!60!black, thick, dashdotted] plot coordinates {(0, 0.157952569) (1, 0.16337234) (2, 0.160156612) (3, 0.164957934) (4, 0.165800938) (5, 0.170247462) (6, 0.165452448) (7, 0.160557051) (8, 0.163440393) (9, 0.163484736) (10, 0.164501886) (11, 0.16522983) (12, 0.165229097) (13, 0.165229108) (14, 0.165229108) (15, 0.165229108) (16, 0.165248094) (17, 0.165242558) (18, 0.165257065) (19, 0.165193481) (20, 0.48753349) (21, 0.520973516) (22, 0.592459502) (23, 0.63283089) (24, 0.660267196) (25, 0.594672238) (26, 0.611962957) (27, 0.824329607) (28, 0.959364766) (29, 0.954862068) (30, 0.973409025) (31, 0.977523414) (32, 0.97760034) (33, 0.97763294) (34, 0.977632997) (35, 0.977528764) (36, 0.977669092) (37, 0.97749226) (38, 0.977674531) (39, 0.977481197) (40, 0.194137836) (41, 0.162985662) (42, 0.173115346) (43, 0.160084783) (44, 0.163555117) (45, 0.163484748) (46, 0.16348475) (47, 0.16348475) (48, 0.16348475) (49, 0.16348475) (50, 0.164501886) (51, 0.16522983) (52, 0.165229097) (53, 0.165229108) (54, 0.165229108) (55, 0.165229108) (56, 0.165248094) (57, 0.165242558) (58, 0.165257065) (59, 0.165193481) (60, 0.48753349) (61, 0.520973516) (62, 0.592459502) (63, 0.63283089) (64, 0.660267196) (65, 0.594672238) (66, 0.611962957) (67, 0.824329607) (68, 0.959364766) (69, 0.954862068) (70, 0.973409025) (71, 0.977523414) (72, 0.97760034) (73, 0.97763294) (74, 0.977632997) (75, 0.977528764) (76, 0.977669092) (77, 0.97749226) (78, 0.977674531) (79, 0.977481197) (80, 0.194137836) (81, 0.162985662) (82, 0.173115346) (83, 0.160084783) (84, 0.163555117) (85, 0.163484748) (86, 0.16348475) (87, 0.16348475) (88, 0.16348475) (89, 0.16348475) (90, 0.164501886) (91, 0.16522983) (92, 0.165229097) (93, 0.165229108) (94, 0.165229108) (95, 0.165229108) (96, 0.165248094) (97, 0.165242558) (98, 0.165257065) (99, 0.165193481)};

        \addplot[blue!60!black, thick, dashed] plot coordinates {(0, 0.921329834) (1, 0.921329834) (2, 0.921329834) (3, 0.921329834) (4, 0.921329834) (5, 0.921329834) (6, 0.921329834) (7, 0.921329834) (8, 0.921329834) (9, 0.921329834) (10, 0.975111457) (11, 0.977457874) (12, 0.977456483) (13, 0.977456504) (14, 0.977456504) (15, 0.977463635) (16, 0.97746502) (17, 0.977463851) (18, 0.977623167) (19, 0.977484889) (20, 0.152518028) (21, 0.152278) (22, 0.150381047) (23, 0.149835799) (24, 0.15247335) (25, 0.150940162) (26, 0.149544233) (27, 0.149544233) (28, 0.149544233) (29, 0.149544233) (30, 0.149622717) (31, 0.149555937) (32, 0.149591943) (33, 0.149590324) (34, 0.149590321) (35, 0.149546682) (36, 0.149581439) (37, 0.149561278) (38, 0.149580289) (39, 0.152775354) (40, 0.921329834) (41, 0.921329834) (42, 0.921329834) (43, 0.921329834) (44, 0.921329834) (45, 0.921329834) (46, 0.921329834) (47, 0.921329834) (48, 0.921329834) (49, 0.921329834) (50, 0.975111457) (51, 0.977457874) (52, 0.977456483) (53, 0.977456504) (54, 0.977456504) (55, 0.977463635) (56, 0.97746502) (57, 0.977463851) (58, 0.977623167) (59, 0.977484889) (60, 0.152518028) (61, 0.152278) (62, 0.150381047) (63, 0.149835799) (64, 0.15247335) (65, 0.150940162) (66, 0.149544233) (67, 0.149544233) (68, 0.149544233) (69, 0.149544233) (70, 0.149622717) (71, 0.149555937) (72, 0.149591943) (73, 0.149590324) (74, 0.149590321) (75, 0.149546682) (76, 0.149581439) (77, 0.149561278) (78, 0.149580289) (79, 0.152775354) (80, 0.921329834) (81, 0.921329834) (82, 0.921329834) (83, 0.921329834) (84, 0.921329834) (85, 0.921329834) (86, 0.921329834) (87, 0.921329834) (88, 0.921329834) (89, 0.921329834) (90, 0.975111457) (91, 0.977457874) (92, 0.977456483) (93, 0.977456504) (94, 0.977456504) (95, 0.977463635) (96, 0.97746502) (97, 0.977463851) (98, 0.977623167) (99, 0.977484889)};
        \addplot[red!60!black, thick, dashed] plot coordinates {(0, 0.149243945) (1, 0.14947931) (2, 0.149084637) (3, 0.149603568) (4, 0.149513471) (5, 0.149414163) (6, 0.14951745) (7, 0.149648871) (8, 0.149648871) (9, 0.149648871) (10, 0.149758051) (11, 0.149791451) (12, 0.149790463) (13, 0.149790462) (14, 0.149790462) (15, 0.149790462) (16, 0.149849311) (17, 0.149843571) (18, 0.14986501) (19, 0.148834106) (20, 0.921438283) (21, 0.921438283) (22, 0.921438283) (23, 0.921438283) (24, 0.921438283) (25, 0.921438283) (26, 0.921438283) (27, 0.921438283) (28, 0.921438283) (29, 0.921438283) (30, 0.97372137) (31, 0.97664628) (32, 0.976671831) (33, 0.976659745) (34, 0.976659697) (35, 0.976477077) (36, 0.97667565) (37, 0.976511595) (38, 0.976686401) (39, 0.976763888) (40, 0.15056309) (41, 0.149879332) (42, 0.149449967) (43, 0.149648871) (44, 0.149648871) (45, 0.149648871) (46, 0.149648871) (47, 0.149648871) (48, 0.149648871) (49, 0.149648871) (50, 0.149758051) (51, 0.149791451) (52, 0.149790463) (53, 0.149790462) (54, 0.149790462) (55, 0.149790462) (56, 0.149849311) (57, 0.149843571) (58, 0.14986501) (59, 0.148834106) (60, 0.921438283) (61, 0.921438283) (62, 0.921438283) (63, 0.921438283) (64, 0.921438283) (65, 0.921438283) (66, 0.921438283) (67, 0.921438283) (68, 0.921438283) (69, 0.921438283) (70, 0.97372137) (71, 0.97664628) (72, 0.976671831) (73, 0.976659745) (74, 0.976659697) (75, 0.976477077) (76, 0.97667565) (77, 0.976511595) (78, 0.976686401) (79, 0.976763888) (80, 0.15056309) (81, 0.149879332) (82, 0.149449967) (83, 0.149648871) (84, 0.149648871) (85, 0.149648871) (86, 0.149648871) (87, 0.149648871) (88, 0.149648871) (89, 0.149648871) (90, 0.149758051) (91, 0.149791451) (92, 0.149790463) (93, 0.149790462) (94, 0.149790462) (95, 0.149790462) (96, 0.149849311) (97, 0.149843571) (98, 0.14986501) (99, 0.149986433)};

        \draw[fill=blue!60!black, draw=none] (0,1.05) rectangle (10,1.12);
        \draw[fill=red!60!black, draw=none] (20,1.05) rectangle (30,1.12);
        \draw[fill=blue!60!black, draw=none] (40,1.05) rectangle (50,1.12);
        \draw[fill=red!60!black, draw=none] (60,1.05) rectangle (70,1.12);
        \draw[fill=blue!60!black, draw=none] (80,1.05) rectangle (90,1.12);
        \draw[black] (\pgfkeysvalueof{/pgfplots/xmin},1.05) rectangle (\pgfkeysvalueof{/pgfplots/xmax},1.12);
        \end{axis}
    \end{tikzpicture}
    \caption{\label{fig:switch}%
    Switching between two memorized periodic firing scores $\breve{x}_B$ (blue) and $\breve{x}_R$ (red).
    The bar at the top shows when and how the network is prompted by 
    $\alpha=75\%$ forced neurons.
    The precision or recall (whichever is smaller)
    with respect to $\breve{x}_B$ and $\breve{x}_R$
    are shown for each period and for three groups of neurons: 
    forceable, non-forceable (autonomous), and all together.
    The plot shows a single (generic) simulation run with
    prompt jitter parameter $\sigma_{\mathrm{s}} / \taumin = 0.05$.
    The distance between the tickmarks is~$10 T$. 
    }
\end{figure}

Numerical results of such an experiment with $\alpha=0.75$
are shown in \figref{fig:switch}. 
Note that the network is able to switch between 
the two memorized attractors, with high precision and recall.

\section{Conclusion}
\label{sec:Conclusion}

We have demonstrated (with numerical experiments) 
that continuous-time recurrent neural networks with random (but fixed) transmission delays 
can store and auto\-no\-mously
reproduce any given random spike trains up to some maximal length $T_\mathrm{max}$, 
with stable accurate relative timing of all spikes, 
with probability close to one.
Moreover, these experiments suggest that 
$T_\mathrm{max}$ scales at least linearly with the number of synaptic inputs per neuron.

For example, we simulated networks with 200 neurons 
(re-)producing tens of firings per neuron (thousands of firings in total)
with accurate relative timing of almost all firings, despite significant threshold noise.
By contrast, previous investigations of precisely timed firings 
were mostly confined to much shorter snippets. 

In these experiments, the required synaptic weights are computed offline
to satisfy a template that encourages temporal stability.
A key new ingredient to this template is the minimum-slope condition~(\ref{eqn:TemplateCondSlope}),
which is shown to be necessary for temporal stability.

Concerning the technical details, we used a new model for 
threshold noise that is easy to simulate and simultaneously models 
jittering, missing, and extra firings.
The simulation of the network used the method proposed by \cite{ComsaAl22}.
The computation of the synaptic weights 
also exploits the special form of the kernel \eqref{eqn:ImpulseResponse}
for efficiency.

As a final remark, 
our experiments demonstrate that clockless continuous-time networks 
can operate with spike-level temporal stability almost like digital processors.
We do not know if this possibility is used in biological neural networks,
but it does seem to open interesting perspectives for neuromorphic engineering.

\section*{Acknowledgment}

The paper builds on the foundations laid by Patrick Murer \citep{Murer22},
who also helped to get the present work started.
We also wish to thank Hampus Malmberg 
for guidance and support.

\appendix

\renewcommand{\thesection}{Appendix~\Alph{section}}
\renewcommand{\theequation}{\Alph{section}.\arabic{equation}}

\section{On the Network Simulations}
\label{app:NetworkSimulation}

\setcounter{equation}{0}

The software for the network simulations in this paper%
\sfootnote{\url{https://github.com/haguettaz/rusty-snn}} 
combines standard ideas of 
event based simulation \citep{PaugamMoisyBohte12}
with the insight by \cite{ComsaAl22} 
concerning the special form of \eqref{eqn:ImpulseResponse}.
For the convenience of the reader 
and in preparation for \ref{app:Optimization}, 
we here outline the main points.

An efficient event-based simulation 
of a recurrent network of spiking neurons typically 
consists of iterating over the following two steps:
\begin{enumerate}
    \item For every neuron in the network, determine the time of the next firing based on accepted past firings.
    This can be done in parallel for all neurons simultaneously.
    \item Accept the earliest of these firings among all neurons and, optionally, all subsequent independent firings
        (i.e., firings that cannot influence each other across all transmission paths).
\end{enumerate}

\noindent
In the following, we focus on Step~1 for a single neuron.
At this point, we do not yet use the special form \eqref{eqn:ImpulseResponse}
of the neural kernel, but we assume it to be continuous, causal, and unimodal.
It follows that the neuron's potential $z(t)$ is continuous.
We aim to determine the next firing time of the neuron, if any, i.e., to compute
\begin{equation} \label{eqn:NextFiringTime}
    t^* \triangleq \min \mleft\{ t \geq t_0 \, | \, z(t) \geq \theta \mright\},
\end{equation}
if such a time exists.
The case $z(t_0) \geq \theta$ trivially yields $t^* = t_0$.
For $z(t_0) < \theta$, 
$t^*$ is
the first time $t > t_0$ where the potential crosses the firing threshold from below.
Let $s_{i}$ denote the $i$th (sorted and delayed) afferent spike of the neuron and $w_{i}$ its associated synaptic weight.
Let $z_i(t)$ be the partial potential
\begin{equation} \label{eqn:PartPotential}
 z_i(t) \triangleq \sum_{j \leq i} w_{j} h(t - s_{j}),
\end{equation}
which satisfies $z_i(t) = z(t)$ for $s_i \leq t < s_{i+1}$.
Then $t^*$ can be computed by repeating the two following steps for $i = 0, 1, \ldots$
{\renewcommand{\theenumi}{1.\arabic{enumi}}%
\begin{enumerate}
    \item \label{enum:StepOneOne}
    Compute the first threshold crossing time $t_i^*$ of $z_i$, 
    i.e., such that $z_i(t_i^*) = \theta$ and $\dot z_i(t_i^*) \geq 0$, if it exists. 
    \item If $s_i \leq t_i^*  < s_{i+1}$ and $t_i^* > t_0$, set $t^* = t_i^*$ and stop. 
    Otherwise, return to step~\ref{enum:StepOneOne} until there are no more afferent spikes.
\end{enumerate}}
\noindent
The process is illustrated in \figref{fig:ThresholdCrossing}.

\begin{figure}[t]
    \centering
    \begin{tikzpicture}
        \begin{groupplot}[group style={group size=4 by 1, horizontal sep=3mm,}, height=4cm, width=4.5cm, xlabel={$t$},
            xmin=0, xmax=3,
            ymin=0.3, ymax=1.4,
            xtick=\empty,
            ytick=\empty,
            clip=false
            ]

            \nextgroupplot[title={$i=0$}]
            \node[label={left:$\theta$}] at (\pgfkeysvalueof{/pgfplots/xmin}, 1) {};
            \draw[dashed] (\pgfkeysvalueof{/pgfplots/xmin}, 1.0) -- (\pgfkeysvalueof{/pgfplots/xmax}, 1.0);
            \addplot[blue!80!black, opacity=0.3, samples=250, domain=0:3] plot {0.75*(x > -0.5) * (x + 0.5) * exp(1-(x + 0.5)) + 0.75*(x > 1) * (x - 1) * exp(1-(x - 1)) - 0.5*(x > 1.125) * (x - 1.125) * exp(1-(x - 1.125)) + 0.75*(x > 1.5) * (x - 1.5) * exp(1-(x - 1.5))};
            \addplot[blue!80!black, thick, samples=250, domain=0:1] plot {0.75*(x > -0.5) * (x + 0.5) * exp(1-(x + 0.5))};

            \nextgroupplot[title={$i=1$}]
            \draw[dashed] (\pgfkeysvalueof{/pgfplots/xmin}, 1.0) -- (\pgfkeysvalueof{/pgfplots/xmax}, 1.0);
            \addplot[blue!80!black, opacity=0.3, samples=250, domain=0:3] plot {0.75*(x > -0.5) * (x + 0.5) * exp(1-(x + 0.5)) + 0.75*(x > 1) * (x - 1) * exp(1-(x - 1)) - 0.5*(x > 1.125) * (x - 1.125) * exp(1-(x - 1.125)) + 0.75*(x > 1.5) * (x - 1.5) * exp(1-(x - 1.5))};
            \addplot[blue!80!black, thick, samples=250, domain=1:1.125] plot {0.75*(x > -0.5) * (x + 0.5) * exp(1-(x + 0.5)) + 0.75*(x > 1) * (x - 1) * exp(1-(x - 1))};

            \nextgroupplot[title={$i=2$}]
            \draw[dashed] (\pgfkeysvalueof{/pgfplots/xmin}, 1.0) -- (\pgfkeysvalueof{/pgfplots/xmax}, 1.0);
            \addplot[blue!80!black, opacity=0.3, samples=250, domain=0:3] plot {0.75*(x > -0.5) * (x + 0.5) * exp(1-(x + 0.5)) + 0.75*(x > 1) * (x - 1) * exp(1-(x - 1)) - 0.5*(x > 1.125) * (x - 1.125) * exp(1-(x - 1.125)) + 0.75*(x > 1.5) * (x - 1.5) * exp(1-(x - 1.5))};            \addplot[blue!80!black, thick, samples=250, domain=1.125:1.5] plot {0.75*(x > -0.5) * (x + 0.5) * exp(1-(x + 0.5)) + 0.75*(x > 1) * (x - 1) * exp(1-(x - 1)) - 0.5*(x > 1.125) * (x - 1.125) * exp(1-(x - 1.125))};

            \nextgroupplot[title={$i=3$}]
            \draw[dashed] (\pgfkeysvalueof{/pgfplots/xmin}, 1.0) -- (\pgfkeysvalueof{/pgfplots/xmax}, 1.0);
            \draw[blue!80!black, dashed] (1.61575,\pgfkeysvalueof{/pgfplots/ymin}) -- (1.61575,\pgfkeysvalueof{/pgfplots/ymax});
            \addplot[blue!80!black, opacity=0.3, samples=250, domain=0:3] plot {0.75*(x > -0.5) * (x + 0.5) * exp(1-(x + 0.5)) + 0.75*(x > 1) * (x - 1) * exp(1-(x - 1)) - 0.5*(x > 1.125) * (x - 1.125) * exp(1-(x - 1.125)) + 0.75*(x > 1.5) * (x - 1.5) * exp(1-(x - 1.5))};
            \addplot[blue!80!black, thick, samples=250, domain=1.5:3] plot {0.75*(x > -0.5) * (x + 0.5) * exp(1-(x + 0.5)) + 0.75*(x > 1) * (x - 1) * exp(1-(x - 1)) - 0.5*(x > 1.125) * (x - 1.125) * exp(1-(x - 1.125)) + 0.75*(x > 1.5) * (x - 1.5) * exp(1-(x - 1.5))};
        \end{groupplot}
    \end{tikzpicture}
    \caption{\label{fig:ThresholdCrossing}%
    Determining the next threshold crossing time of the neuron potential. 
    The partial potential \eqref{eqn:PartPotential}
    is shown (in bold blue) in the interval of its validity;
    the actual potential is shown in faint blue.
    }
\end{figure}

Because $z_i(t)$ is continuous, step~\ref{enum:StepOneOne} can be numerically solved by bracketing.
However, 
the special form of \eqref{eqn:ImpulseResponse} (known as alpha kernel)
allows a much more efficient way,
as was noted by \cite{ComsaAl22}.
Plugging \eqref{eqn:ImpulseResponse} into \eqref{eqn:PartPotential}, we obtain
\begin{IEEEeqnarray}{rCl}
 z_i(t) 
    &=& \sum_{j \leq i} w_j (t - s_j) e^{1 - (t - s_j)} \\
    &=& \sum_{j \leq i} w_j (t - s_j) e^{1 - (s_i - s_j) - (t - s_i)}  \IEEEeqnarraynumspace\\
    &=& (a_i t - b_i ) \, e^{- (t - s_i)}  \label{eqn:PartPotentialAlpha}
    \IEEEeqnarraynumspace
\end{IEEEeqnarray}
with
\begin{equation} \label{eqn:AlphaDefai}
a_i \triangleq \sum_{j \leq i} w_j e^{1 - (s_i - s_j)}
\end{equation}
and
\begin{equation} \label{eqn:AlphaDefbi}
b_i \triangleq \sum_{j \leq i} s_j w_j e^{1 - (s_i - s_j)}.
\end{equation}
Solving $z_i(t_i) = \theta$ for $t_i \in [s_i, s_{i+1})$ we obtain
\begin{equation}
 t_i = 
    \begin{cases}
 s_i - \ln(- \theta / b_i) & \text{if it is valid and } a_i = 0, \\
 b_i / a_i - W_0(- \theta \, e^{b_i / a_i - s_i } \, / a_i) & \text{if it is valid and } a_i \neq 0, \\
 \text{undefined} & \text{otherwise,}
    \end{cases}
\end{equation}
where $W_0$ denotes the first branch of the Lambert function,
and ``valid'' means that the expression is well defined and yields $t_i$ in the interval $[s_i, s_{i+1})$.
Note that the quantities 
\eqref{eqn:AlphaDefai} and \eqref{eqn:AlphaDefbi}
can and should be computed recursively.

Finally, we note also that the potential $z(t)$ at any time $t$ 
can be easily computed with \eqref{eqn:PartPotentialAlpha}
after having identified 
the pertinent interval $[s_i, s_{i+1}) \ni t$.

\section{Computing the Synaptic Weights by Iterative Constraint Refinement}
\label{app:Optimization}

\setcounter{equation}{0}

As stated in Section~\ref{sec:compute_weights},
the synaptic weights are determined by minimizing the quadratic norm \eqref{eqn:RegL2} 
subject to the constraints \eqref{eqn:TemplateCondCrossing}--\eqref{eqn:WeightsBound}.
As noted in Section~\ref{sec:compute_weights},
this optimization problem splits into independent optimizations for each neuron.
Dropping the neuron index $\ell$, 
the continuous-time constraints \eqref{eqn:TemplateCondCrossing}--\eqref{eqn:TemplateCondSlope}
are
\begin{IEEEeqnarray}{CCl}
    z(t) = \theta_0  & \qquad & \text{if $t\in\breve{\set{S}}$} \label{eqn:AppTemplateCondCrossing}\\
    z(t) <  \zmax & & \text{unless $s-\varepsilon_{\mathrm{s}} < t<s+\taumin$ for some $s\in \breve{\set{S}}$} \label{eqn:AppTemplateCondExtraBelow} \\
    \dot z(t) > \dzmin  & & \text{if $s-\varepsilon_{\mathrm{s}} < t<s+\varepsilon_{\mathrm{s}}$ for some $s\in \breve{\set{S}}$}. \label{eqn:AppTemplateCondSlope}
\end{IEEEeqnarray}

In this paper, we address this optimization problem
by converting \eqref{eqn:AppTemplateCondExtraBelow} and~\eqref{eqn:AppTemplateCondSlope}
into an equivalent list of constraints for discrete points in time. 
This list of constraints is constructed by the following iterative algorithm.
\begin{enumerate}
    \item 
    Initialize the list with the firing time constraints \eqref{eqn:AppTemplateCondCrossing}.
    \item
    Compute the optimal synaptic weights according to the momentary list of constraints.
    (This can be done by standard optimization software.)
    \item 
    For fixed synaptic weights, 
    compute 
    \begin{equation} \label{eqn:WeightArgmaxPot}
      \hat t \triangleq \argmax_{t} z(t)
    \end{equation}
    for every interval according to \eqref{eqn:AppTemplateCondExtraBelow};
    if $z(\hat t) \geq \zmax$, 
    then add the constraint $z(\hat t) < \zmax$ to the list.

    Likewise, compute
    \begin{equation} \label{eqn:WeightArginSlope}
      \breve t \triangleq \argmin_{t} \dot z(t)
    \end{equation}
    for every interval according to \eqref{eqn:AppTemplateCondSlope};
    if $z(\breve t) \leq \dzmin$, 
    then add the constraint $z(\breve t) > \dzmin$ to the list.
    \item
    If the list was modified (i.e., extended) in step~3, then go to step~2;
    otherwise, stop.
    \end{enumerate}

When this algorithm terminates, its final synaptic weights are the solution of the original optimization problem.
(We do not have any theoretical guarantees for termination, but it always seems to terminate in practice.)

\begin{figure}[t]
    \centering
    \begin{tikzpicture}
        \begin{groupplot}[group style={group size=4 by 1, horizontal sep=3mm}, height=4cm, width=4.5cm, xlabel={$t$},
            xmin=0, xmax=3,
            ymin=0.3, ymax=1.4,
            xtick=\empty,
            ytick=\empty
            ]

            \nextgroupplot[title={$i=0$}]
            \draw[dashed, blue!80!black] (0.5,\pgfkeysvalueof{/pgfplots/ymin}) -- (0.5,\pgfkeysvalueof{/pgfplots/ymax});
            \addplot[blue!80!black, opacity=0.3, samples=250, domain=0:3] plot {0.75*(x > -0.5) * (x + 0.5) * exp(1-(x + 0.5)) + 0.75*(x > 1) * (x - 1) * exp(1-(x - 1)) - 0.5*(x > 1.125) * (x - 1.125) * exp(1-(x - 1.125)) + 0.75*(x > 1.5) * (x - 1.5) * exp(1-(x - 1.5))};
            \addplot[blue!80!black, thick, samples=250, domain=0:1] plot {0.75*(x > -0.5) * (x + 0.5) * exp(1-(x + 0.5))};

            \nextgroupplot[title={$i=1$}]
            \draw[dashed, blue!80!black] (1.125,\pgfkeysvalueof{/pgfplots/ymin}) -- (1.125,\pgfkeysvalueof{/pgfplots/ymax});
            \addplot[blue!80!black, opacity=0.3, samples=250, domain=0:3] plot {0.75*(x > -0.5) * (x + 0.5) * exp(1-(x + 0.5)) + 0.75*(x > 1) * (x - 1) * exp(1-(x - 1)) - 0.5*(x > 1.125) * (x - 1.125) * exp(1-(x - 1.125)) + 0.75*(x > 1.5) * (x - 1.5) * exp(1-(x - 1.5))};
            \addplot[blue!80!black, thick, samples=250, domain=1:1.125] plot {0.75*(x > -0.5) * (x + 0.5) * exp(1-(x + 0.5)) + 0.75*(x > 1) * (x - 1) * exp(1-(x - 1))};

            \nextgroupplot[title={$i=2$}]
            \draw[dashed, blue!80!black] (1.125,\pgfkeysvalueof{/pgfplots/ymin}) -- (1.125,\pgfkeysvalueof{/pgfplots/ymax});
            \addplot[blue!80!black, opacity=0.3, samples=250, domain=0:3] plot {0.75*(x > -0.5) * (x + 0.5) * exp(1-(x + 0.5)) + 0.75*(x > 1) * (x - 1) * exp(1-(x - 1)) - 0.5*(x > 1.125) * (x - 1.125) * exp(1-(x - 1.125)) + 0.75*(x > 1.5) * (x - 1.5) * exp(1-(x - 1.5))};
            \addplot[blue!80!black, thick, samples=250, domain=1.125:1.5] plot {0.75*(x > -0.5) * (x + 0.5) * exp(1-(x + 0.5)) + 0.75*(x > 1) * (x - 1) * exp(1-(x - 1)) - 0.5*(x > 1.125) * (x - 1.125) * exp(1-(x - 1.125))};

            \nextgroupplot[title={$i=3$}]
            \draw[dashed, blue!80!black] (2.18675,\pgfkeysvalueof{/pgfplots/ymin}) -- (2.18675,\pgfkeysvalueof{/pgfplots/ymax});
            \addplot[blue!80!black, opacity=0.3, samples=250, domain=0:3] plot {0.75*(x > -0.5) * (x + 0.5) * exp(1-(x + 0.5)) + 0.75*(x > 1) * (x - 1) * exp(1-(x - 1)) - 0.5*(x > 1.125) * (x - 1.125) * exp(1-(x - 1.125)) + 0.75*(x > 1.5) * (x - 1.5) * exp(1-(x - 1.5))};
            \addplot[blue!80!black, thick, samples=250, domain=1.5:3] plot {0.75*(x > -0.5) * (x + 0.5) * exp(1-(x + 0.5)) + 0.75*(x > 1) * (x - 1) * exp(1-(x - 1)) - 0.5*(x > 1.125) * (x - 1.125) * exp(1-(x - 1.125)) + 0.75*(x > 1.5) * (x - 1.5) * exp(1-(x - 1.5))};
        \end{groupplot}
    \end{tikzpicture}
    \caption{\label{fig:Maximization}%
    Finding the maximum of the neuron potential over a closed interval. 
    The partial potential \eqref{eqn:PartPotentialAlpha} is shown (in bold blue) 
    in the interval of its validity; the actual potential is shown in faint blue.
    At every step, the so-far maximizer is indicated by a dashed line.}
\end{figure}

For the specific neural kernel \eqref{eqn:ImpulseResponse},
we can use \eqref{eqn:PartPotentialAlpha} 
to compute \eqref{eqn:WeightArgmaxPot} and \eqref{eqn:WeightArginSlope}
analytically.
Concerning \eqref{eqn:WeightArgmaxPot}, 
the local maxima of $z(t)$ are found among the following points:
\begin{enumerate}
    \item the non-differentiable points $s_i$ and $s_{i+1}$,
    \item the boundaries of the interval of interest, if applicable,
    \item the zero-slope points with negative second derivative.
\end{enumerate}
The resulting algorithm is illustrated in \figref{fig:Maximization}. 

Concerning the analytical maximization, 
recall the coefficients $a_i$ and $b_i$ in \eqref{eqn:PartPotentialAlpha}.

If $a_i\neq 0$, then $\dot z_i(t)=0$ if and only if 
\begin{equation} \label{eqn:OptimMaxSaddle}
    t = 1 + b_i / a_i,
\end{equation}
which is a (local) maximum if and only if $a_i>0$.
If $a_i=0$ and $b_i < 0$, then \eqref{eqn:PartPotentialAlpha} is
decreasing; hence a local maximum is attained at $t_i = s_i$.
If $a_i=0$ and $b_i > 0$, then \eqref{eqn:PartPotentialAlpha} is
increasing; hence a local maximum is attained at $t_i = s_{i+1}$.

The computation of \eqref{eqn:WeightArginSlope}
is very similar. In particular, condition \eqref{eqn:OptimMaxSaddle} is replaced by
\begin{equation} \label{eqn:OptimMinSaddle}
    t = 2 + b_i / a_i,
\end{equation}
which is a local minimum if and only if $a_i>0$.

\section{The Derivatives \eqref{eqn:LinJitterPropCoeffs}}
\label{app:LinJitterProp}

\setcounter{equation}{0}

Recall the setting of \eqref{eqn:LinJitterPropCoeffs},
with nominal spike times 
$\breve s_0, \breve s_1,\, \ldots$ 
(with $\breve s_0 \leq \breve s_1 \leq \ldots$),
actual spike times 
$s_0 = \breve s_0 + ds_0, 
 s_1 = \breve s_1 + ds_1,
 \ldots,$
and no threshold noise.
Let $N$ be the number of spikes in the first period. 

Let $z_{\ell(n)}(t)$ be the potential of the neuron that produces the firing at time $s_n$.
For $n>N$, we can write
\begin{equation} \label{app:eqn:PotentialFiringDecomp}
    z_{\ell(n)}(t) = \sum_{n'=1}^N z_{n,n'}(t), 
\end{equation}
where $z_{n,n'}(t)$ is the contribution to $z_{\ell(n)}(t)$ by the spike at time $s_{n-n'}$.
Let $\breve z_{\ell(n)}(t)$ and $\breve z_{n,n'}(t)$ be the nominal values of
$z_{\ell(n)}(t)$ and $z_{n,n'}(t)$, respectively.
Since there is no threshold noise, we have 
both $z_{\ell(n)}(s_n) = \theta_0$ and $\breve z_{\ell(n)}(\breve s_n) = \theta_0$
for all $n$.
Note also that
\begin{equation}
z_{n,n'}(t) = \breve z_{n,n'}(t - ds_{n-n'}).
\end{equation}

In the following, $\dot{\breve z}_{n,n'}(t)$ denotes the derivative of ${\breve z}_{n,n'}(t)$.
Linearizing \eqref{app:eqn:PotentialFiringDecomp} around the nominal firing time $\breve s_{\ell(n)}$ yields%

\begin{eqnarray}
    z_{\ell(n)}(\breve s_n + ds_n) &=& \sum_{n'} z_{n,n'}(\breve s_n + ds_n) \\
    &=& \sum_{n'} \breve z_{n,n'}(\breve s_n + ds_n - ds_{n-n'})\\ 
    &\approx& \sum_{n'}  \breve z_{n,n'}(\breve s_n) + \sum_{n'} (ds_{n} - ds_{n - n'}) \, \dot{\breve z}_{n,n'}(\breve s_n) \\
    &=& \breve z_{\ell(n)}(\breve s_n) + \sum_{n'} (ds_{n} - ds_{n - n'}) \, \dot{\breve z}_{n,n'}(\breve s_n) \\
    &=& \theta_0 + \sum_{n'} (ds_{n} - ds_{n - n'}) \, \dot{\breve z}_{n,n'}(\breve s_n).
\end{eqnarray}

Solving $z_{\ell(n)}(\breve s_n + ds_n) = \theta_0$ for $ds_n$ then yields
\begin{equation}
    ds_n = \sum_{n'=1}^N a_{n, n'} ds_{n - n'}
\end{equation}
with 
\begin{equation} \label{app:eqn:LinJitterPropCoeffs}
    a_{n, n'} = \frac{\dot{\breve z}_{n,n'}(\breve s_n)}{\sum_{n''} \dot{\breve z}_{n,n''}(\breve s_n)}
\end{equation}
Finally, from \eqref{eqn:PotentialSum}, we have
\begin{equation}
\dot{\breve z}_{n,n'}(\breve s_n) = \sum_{k\in \mathcal{K}(n,n')} w_{\ell(n),k} \, \dot{h}(\breve s_n - d_{\ell(n),k} - \breve s_{n-n'}),
\end{equation}
where $\mathcal{K}(n,n')$ is the set of indices $k$
such that the firing at time $\breve s_{n-n'}$ contributes to $z_{\ell(n)}(t)$ via $\tilde{x}_{\ell(n),k}$.
Note that the nominal spikes being periodic with period $T$, 
we have $\dot{\breve z}_{n + mN,n'}(\breve s_n + mT) = \dot{\breve z}_{n,n'}(\breve s_n)$ 
and thus $a_{n + mN,n'} = a_{n, n'}$ for every $m \geq 0$.

\section{The Random Periodic Spike Trains}
\label{app:SpikeTrain}

\setcounter{equation}{0}
\setcounter{subsection}{0}
\renewcommand{\thesubsection}{\Alph{section}.\arabic{subsection}}

In this section, we provide the definition of,
and the sampling algorithm for, the periodic random spike trains
that are used in this paper
as described in Section~\ref{sec:PrescribedScore}.

As stated in Section~\ref{sec:PrescribedScore},
the random spike trains for the different neurons 
are statistically independent. Therefore, it suffices 
to consider here only a single spike train (i.e., a single neuron).

The basic idea is to modify the definition of a stationary Poisson process 
with firing rate $\lambda$
such that
\begin{itemize}
\item
it is periodic (with period $T$), and
\item
it respects the refractory period, i.e., any two spikes are separated 
by at least $\taumin$.
\end{itemize}
The pertinent math is given 
in Sections \ref{app:Poisson} and~\ref{app:PoissonModified}.
The resulting sampling algorithm is given in Section~\ref{sec:SamplingAlg}.
The expected number of spikes per period is given by \eqref{app:eqn:ExpNumSpikes} below.

\subsection{Poisson Process in an Interval}
\label{app:Poisson}

In a (standard stationary) Poisson process with firing rate $\lambda>0$, 
the number of firings $N$ in an interval of duration $T$ is a random variable,
and the probability of exactly $n$ firings is
\begin{equation} \label{eqn:Poisson}
P(N=n) = \gamma_\text{P} \frac{(\lambda T)^n}{n!}
\end{equation}
with scale factor $\gamma_\text{P}=\exp(-\lambda T)$.
In order to suitably modify (\ref{eqn:Poisson}) for our purpose, 
we need to dig a little deeper. 
Divide $T$ into $K$ subintervals of length $T/K$ and let $N_K$ be the (random)
number of firings in some fixed subinterval.
For large~$K$, 
each subinterval contains at most one firing, i.e., 
\begin{equation} \label{eqn:PoissonSmallInverval}
P(N_K > 1) \ll P(N_K = 1) \approx \lambda T/K
\end{equation}
(which becomes exact in the limit $K\rightarrow\infty$),
and \eqref{eqn:Poisson} can be recovered from (\ref{eqn:PoissonSmallInverval}) by
\begin{equation} \label{eqn:PoissonDecomp}
P(N=n) = \lim_{K \to \infty} 
         \binom{K}{n} \left( \frac{\lambda T}{K} \right)^n
         \left( 1 - \frac{\lambda T}{K} \right)^{K-n}.
\end{equation}

\subsection{Periodic Poisson Spike Trains}
\label{app:PoissonModified}

We keep the probability measure of the Poisson process over $[0,T)$
except that we forbid all configurations that violate the refractory period.
Specifically, we replace \eqref{eqn:PoissonDecomp} by
\begin{equation} \label{eqn:ModifiedPoissonStart}
P(N=n) = \tilde\gamma \lim_{K\rightarrow\infty} 
         C(K, n)
         \mleft( \frac{\lambda T}{K} \mright)^n
         \mleft( 1 - \frac{\lambda T}{K} \mright)^{K-n},
\end{equation}
where $C(K,n)$ is the number of configurations 
(with at most one firing per subinterval and $n$ firings in total)
with at least $K \taumin/T$ empty subintervals between firings.
The scale factor $\tilde\gamma$ 
is required for (\ref{eqn:ModifiedPoissonStart}) to be properly normalized.

In order to determine $C(K,n)$, 
we note that there is a one-to-one correspondence 
(illustrated in \figref{fig:PoissonProcess}) 
between
\begin{itemize}
\item[(a)]
configurations of $n$ spikes on the interval $[0,T)$ such that (i) there is a spike at~0
and (ii) each spike is followed by a refractory period of length $\taumin$,
and
\item[(b)]
configurations of $n$ spikes on the interval $[0, T - n \taumin)$ with a spike at~0.
\end{itemize}

Counting the latter is obvious: there are 
\begin{equation}
(K_n - 1) (K_n - 2) \cdots (K_n - n+1)
\end{equation}
such ordered configurations where 
\begin{equation}
K_n \triangleq \mleft\lfloor K \mleft( 1 - \frac{n\taumin}{T} \mright) \mright\rfloor.
\end{equation}
Dropping the constraint that the first spike occurs at time~0
and passing to unordered configurations yields
\begin{equation}
C(K,n) = \frac{K (K_n - 1) (K_n - 2) \cdots (K_n - n+1)}{n!}
\end{equation}
with $C(K,0)=1$.
For $n>0$, we have
\begin{eqnarray}
P(N=n) & = & \tilde\gamma \lim_{K\rightarrow\infty} 
         C(K, n)
         \mleft( \frac{\lambda T/K}{1 - \lambda T/K} \mright)^n
         \mleft( 1 - \frac{\lambda T}{K} \mright)^K, \\
& = &  \tilde\gamma \lim_{K\rightarrow\infty} 
        C(K, n)
        \mleft( \frac{\lambda T}{K} \mright)^n e^{-\lambda T} \\
& = &  \tilde\gamma \frac{\mleft( 1 - \frac{n\taumin}{T} \mright)^{n-1}}{n!} 
             (\lambda T)^n e^{-\lambda T} \\
& = &  \tilde\gamma \frac{\big( \lambda (T - n\taumin) \big)^{n-1}}{n!} 
              \lambda T e^{-\lambda T},           
        \label{eqn:EndProfProb}
\end{eqnarray}
which happens to hold also for $n=0$.
We thus have
\begin{equation} \label{eqn:PMFNumSpikes}
P(N=n) = 
    \begin{dcases}
        \gamma \dfrac{\big(\lambda(T - n \taumin)\big)^{n-1}}{n!}, & \text{if $0 \leq n < T/\taumin$} \\
        0,  & \text{if $n \geq T/\taumin$,}
    \end{dcases}
\end{equation}
where the scale factor $\gamma$ is determined by $\sum_n P(N=n) = 1$.
Note that
$P(N\! =\! 0)=\gamma(\lambda T)^{-1}$.
Note also that 
\eqref{eqn:PMFNumSpikes} agrees with \eqref{eqn:Poisson}
for $\taumin=0$ and $\gamma = \gamma_\text{P} \lambda T$.

The expected number of spikes per period is
\begin{equation} \label{app:eqn:ExpNumSpikes}
\E{N} = \sum_n n P(N=n).
\end{equation}
Numerical examples of \eqref{eqn:PMFNumSpikes} and \eqref{app:eqn:ExpNumSpikes}
are given in \figref{fig:PMFNumSpikes}.

\begin{figure}[t]
    \centering
    \begin{tikzpicture}[>=latex]
        \tikzset{dot/.style={draw,blue!80!black,circle,fill=blue!80!black, minimum size=2mm, inner sep=0pt}}

        \fill[blue!10] (0,1) -- (0,0) -- (1,0) -- cycle;
        \fill[blue!10] (1.6,1) -- (2.6,0) -- (3.6,0) -- cycle;
        \fill[blue!10] (2.6,1) -- (4.6,0) -- (5.6,0) -- cycle;

        \node[dot, label={below:$s_0$}] (s0) at (0,0) {};
        \node[dot, label={below:$s_1$}] (s1) at (2.6,0) {};
        \node[dot, label={below:$s_2$}] (s2) at (4.6,0) {};
        \node[dot, black, fill=white] (s3) at (8,0) {};
        
        \draw (s0) -- (s1) -- (s2) -- (s3);

        \draw[very thick, blue!80!black] (0,0) -- (1,0);
        \draw[very thick, blue!80!black] (2.6,0) -- (3.6,0);
        \draw[very thick, blue!80!black] (4.6,0) -- (5.6,0);

        \draw[<->] (0,-0.65) -- (8,-0.65) node[midway, below] {$T$};

        \node[dot, label={above:$0$}] (u0) at (0,1) {};
        \node[dot, label={above:$u_1$}] (u1) at (1.6,1) {};
        \node[dot, label={above:$u_2$}] (u2) at (2.6,1) {};
        \node[dot, black, fill=white] (u3) at (5,1) {};

        \draw (u0) -- (u1) -- (u2) -- (u3);

        \draw[<->] (0,1.7) -- (5,1.7) node[midway, above] {$T - n \tau_0$};
    \end{tikzpicture}
    \caption{\label{fig:PoissonProcess}
    The one-to-one correspondence of configurations in Section \ref{app:PoissonModified}.}
\end{figure}

\begin{figure}[t]
    \centering
    \begin{tikzpicture}
        \begin{axis}[
            width=0.9\textwidth,
            height=0.4\textwidth,
            tick pos=left,
            xlabel={$n$},
            ylabel={$P(N=n)$},
            xmin=0, xmax=35,
            ymin=0, ymax=0.23,
            xmajorgrids,
            ymajorgrids,
          ]
      
          \addplot[ycomb, red!80!black, mark=*, thick, mark size=1.4pt] plot coordinates {(0,0.000) (1,0.002) (2,0.010) (3,0.032) (4,0.070) (5,0.117) (6,0.157) (7,0.172) (8,0.157) (9,0.121) (10,0.079) (11,0.045) (12,0.022) (13,0.009) (14,0.003) (15,0.001) (16,0.000) (17,0.000) (18,0.000) (19,0.000) (20,0.000) (21,0.000) (22,0.000) (23,0.000) (24,0.000) (25,0.000) (26,0.000) (27,0.000) (28,0.000) (29,0.000) (30,0.000) (31,0.000) (32,0.000) (33,0.000) (34,0.000) (35,0.000) (36,0.000) (37,0.000) (38,0.000) (39,0.000) (40,0.000) (41,0.000) (42,0.000) (43,0.000) (44,0.000) (45,0.000) (46,0.000) (47,0.000) (48,0.000) (49,0.000)};
          \addplot[red!80!black, thick, opacity=0.5, forget plot, smooth] plot coordinates {(0,0.000) (1,0.002) (2,0.010) (3,0.032) (4,0.070) (5,0.117) (6,0.157) (7,0.172) (8,0.157) (9,0.121) (10,0.079) (11,0.045) (12,0.022) (13,0.009) (14,0.003) (15,0.001) (16,0.000) (17,0.000) (18,0.000) (19,0.000) (20,0.000) (21,0.000) (22,0.000) (23,0.000) (24,0.000) (25,0.000) (26,0.000) (27,0.000) (28,0.000) (29,0.000) (30,0.000) (31,0.000) (32,0.000) (33,0.000) (34,0.000) (35,0.000) (36,0.000) (37,0.000) (38,0.000) (39,0.000) (40,0.000) (41,0.000) (42,0.000) (43,0.000) (44,0.000) (45,0.000) (46,0.000) (47,0.000) (48,0.000) (49,0.000)};
          
          \addplot[ycomb, green!60!black, mark=*, thick, mark size=1.4pt] plot coordinates {(0,0.000) (1,0.000) (2,0.000) (3,0.000) (4,0.000) (5,0.001) (6,0.004) (7,0.011) (8,0.026) (9,0.049) (10,0.081) (11,0.115) (12,0.140) (13,0.149) (14,0.137) (15,0.111) (16,0.079) (17,0.049) (18,0.027) (19,0.013) (20,0.005) (21,0.002) (22,0.001) (23,0.000) (24,0.000) (25,0.000) (26,0.000) (27,0.000) (28,0.000) (29,0.000) (30,0.000) (31,0.000) (32,0.000) (33,0.000) (34,0.000) (35,0.000) (36,0.000) (37,0.000) (38,0.000) (39,0.000) (40,0.000) (41,0.000) (42,0.000) (43,0.000) (44,0.000) (45,0.000) (46,0.000) (47,0.000) (48,0.000) (49,0.000)};
          \addplot[green!60!black, thick, opacity=0.5, forget plot, smooth] plot coordinates {(0,0.000) (1,0.000) (2,0.000) (3,0.000) (4,0.000) (5,0.001) (6,0.004) (7,0.011) (8,0.026) (9,0.049) (10,0.081) (11,0.115) (12,0.140) (13,0.149) (14,0.137) (15,0.111) (16,0.079) (17,0.049) (18,0.027) (19,0.013) (20,0.005) (21,0.002) (22,0.001) (23,0.000) (24,0.000) (25,0.000) (26,0.000) (27,0.000) (28,0.000) (29,0.000) (30,0.000) (31,0.000) (32,0.000) (33,0.000) (34,0.000) (35,0.000) (36,0.000) (37,0.000) (38,0.000) (39,0.000) (40,0.000) (41,0.000) (42,0.000) (43,0.000) (44,0.000) (45,0.000) (46,0.000) (47,0.000) (48,0.000) (49,0.000)};
      
          \addplot[ycomb, blue!80!black, mark=*, thick, mark size=1.4pt] plot coordinates {(0, 0.0) (1, 0.0) (2, 0.0) (3, 0.0) (4, 0.0) (5, 0.0) (6, 0.0) (7, 0.0) (8, 0.0001) (9, 0.0005) (10, 0.0017) (11, 0.0049) (12, 0.012) (13, 0.0256) (14, 0.0473) (15, 0.0766) (16, 0.1084) (17, 0.1345) (18, 0.1462) (19, 0.139) (20, 0.1156) (21, 0.0838) (22, 0.0529) (23, 0.0289) (24, 0.0137) (25, 0.0055) (26, 0.0019) (27, 0.0006) (28, 0.0001) (29, 0.0) (30, 0.0) (31, 0.0) (32, 0.0) (33, 0.0) (34, 0.0) (35, 0.0) (36, 0.0) (37, 0.0) (38, 0.0) (39, 0.0) (40, 0.0) (41, 0.0) (42, 0.0) (43, 0.0) (44, 0.0) (45, 0.0) (46, 0.0) (47, 0.0) (48, 0.0) (49, 0.0)};
          \addplot[blue!80!black, thick, opacity=0.5, forget plot, smooth] plot coordinates {(0, 0.0) (1, 0.0) (2, 0.0) (3, 0.0) (4, 0.0) (5, 0.0) (6, 0.0) (7, 0.0) (8, 0.0001) (9, 0.0005) (10, 0.0017) (11, 0.0049) (12, 0.012) (13, 0.0256) (14, 0.0473) (15, 0.0766) (16, 0.1084) (17, 0.1345) (18, 0.1462) (19, 0.139) (20, 0.1156) (21, 0.0838) (22, 0.0529) (23, 0.0289) (24, 0.0137) (25, 0.0055) (26, 0.0019) (27, 0.0006) (28, 0.0001) (29, 0.0) (30, 0.0) (31, 0.0) (32, 0.0) (33, 0.0) (34, 0.0) (35, 0.0) (36, 0.0) (37, 0.0) (38, 0.0) (39, 0.0) (40, 0.0) (41, 0.0) (42, 0.0) (43, 0.0) (44, 0.0) (45, 0.0) (46, 0.0) (47, 0.0) (48, 0.0) (49, 0.0)};
          
          \addplot[ycomb, black, mark=*, thick, mark size=1.4pt] plot coordinates {(0, 0.0) (1, 0.0) (2, 0.0) (3, 0.0) (4, 0.0) (5, 0.0) (6, 0.0) (7, 0.0) (8, 0.0) (9, 0.0) (10, 0.0) (11, 0.0) (12, 0.0) (13, 0.0001) (14, 0.0005) (15, 0.0016) (16, 0.0045) (17, 0.0112) (18, 0.0242) (19, 0.0461) (20, 0.0767) (21, 0.1112) (22, 0.1403) (23, 0.1535) (24, 0.145) (25, 0.1176) (26, 0.0815) (27, 0.0479) (28, 0.0237) (29, 0.0098) (30, 0.0033) (31, 0.0009) (32, 0.0002) (33, 0.0) (34, 0.0) (35, 0.0) (36, 0.0) (37, 0.0) (38, 0.0) (39, 0.0) (40, 0.0) (41, 0.0) (42, 0.0) (43, 0.0) (44, 0.0) (45, 0.0) (46, 0.0) (47, 0.0) (48, 0.0) (49, 0.0)};
          \addplot[black, thick, opacity=0.5, forget plot, smooth] plot coordinates {(0, 0.0) (1, 0.0) (2, 0.0) (3, 0.0) (4, 0.0) (5, 0.0) (6, 0.0) (7, 0.0) (8, 0.0) (9, 0.0) (10, 0.0) (11, 0.0) (12, 0.0) (13, 0.0001) (14, 0.0005) (15, 0.0016) (16, 0.0045) (17, 0.0112) (18, 0.0242) (19, 0.0461) (20, 0.0767) (21, 0.1112) (22, 0.1403) (23, 0.1535) (24, 0.145) (25, 0.1176) (26, 0.0815) (27, 0.0479) (28, 0.0237) (29, 0.0098) (30, 0.0033) (31, 0.0009) (32, 0.0002) (33, 0.0) (34, 0.0) (35, 0.0) (36, 0.0) (37, 0.0) (38, 0.0) (39, 0.0) (40, 0.0) (41, 0.0) (42, 0.0) (43, 0.0) (44, 0.0) (45, 0.0) (46, 0.0) (47, 0.0) (48, 0.0) (49, 0.0)};
      
          \draw[thick, red!80!black, dashed] (7.225,0) -- (7.225,1);
          \draw[thick, green!60!black, dashed] (13.010,0) -- (13.010,1);
          \draw[thick, blue!80!black, dashed] (18.095,0) -- (18.095,1);
          \draw[thick, black, dashed] (23.011,0) -- (23.011,1);
      
          \legend{$\lambda \tau_{0}=0.2$,$\lambda \tau_{0}=0.5$,$\lambda \tau_{0}=1.0$,$\lambda \tau_{0}=2.0$}
        \end{axis}
    \end{tikzpicture}
    \caption{The probability mass function \eqref{eqn:PMFNumSpikes} with $T = 50 \, \taumin$ 
    for different values of $\lambda \taumin$. 
    The expected number of spikes \eqref{app:eqn:ExpNumSpikes} 
    is shown as a vertical dashed line.}
    \label{fig:PMFNumSpikes}
\end{figure}

\subsection{The Sampling Algorithm}
\label{sec:SamplingAlg}

We thus arrive at the following sampling algorithm.
\begin{enumerate}
\item
Sample the number of spikes $n$ per period according to \eqref{eqn:PMFNumSpikes}.
If $n=0$, return an empty spike train.
\item
Sample $s_0$, the position of the first spike, uniformly on $[0, T)$.
\item
Sample $u_1, u_2, ..., u_{n-1}$ independently and uniformly on [$0,T-n\taumin]$,
and (with a slight abuse of notation) sort them such that $u_1 < u_2 < \ldots < u_{n-1}$.
\item
For $n'=1,\ldots,n-1$, 
the firing times are $s_{n'} = s_0 + n' \taumin + u_{n'}$
(cf.\ \figref{fig:PoissonProcess}).
\item
Periodically extend $s_0, s_1, ..., s_{n-1}$.
\end{enumerate}

\section{Adding Spike Jitter Respecting the Refractory Period}
\label{appsec:PromptJitter}

\setcounter{equation}{0}

\begin{figure}
    \centering
    \begin{tikzpicture}[node distance=14mm, >=latex]
        \tikzset{box/.style={draw,rectangle,minimum size=6mm, inner sep=0}}
        
        \node[box, dashed, label=below:$g_0$] (g0) at (0,0) {};
        \node[box, right=of g0] (eq1){$=$};
        \node[box, right=of eq1, label=below:$g$] (g1) {};
        \node[box, above=of eq1, label=above:$\Normal{\breve s_0}{\sigma_s^2}$] (p1) {};
        \node[box, right=of g1] (eq2) {$=$};
        \node[box, above=of eq2, label=above:$\Normal{\breve s_1}{\sigma_s^2}$] (p2) {};
        \node[box, right=18mm of eq2] (eq3) {$=$};
        \node[box, above=of eq3, label=above:$\Normal{\breve s_{n-1}}{\sigma_s^2}$] (p3) {};
        \node[box, dashed, right=of eq3, label=below:$g_{n-1}$] (g4) {};
        
        \node (dots) at ($(eq2)!0.5!(eq3)$) {$\dots$};
        
        \draw[->] (g0) -- (eq1);
        \draw[->] (p1) -- (eq1) node[midway, right] {$S_0$};
        \draw[->] (eq1) -- (g1);
        \draw[->] (g1) -- (eq2);
        \draw[->] (p2) -- (eq2) node[midway, right] {$S_1$};
        \draw (eq2) -- (dots) -- (eq3);
        \draw[->] (p3) -- (eq3) node[midway, right] {$S_{n-1}$};
        \draw[->] (eq3) -- (g4);

        \node[above=of p1] {};
    \end{tikzpicture}
    \caption{\label{eqn:JitterFG}%
    Factor graph \citep{Loeliger2004}
    of the joint probability density function of $S_0,\ldots,S_{n-1}$
    with $g \triangleq \mathbbm{1}[S_{n'} - S_{n'-1} > \taumin]$ for $n' = 1, ..., n-1$, 
    and optional additional constraints on $S_0$ and $S_{n-1}$ expressed by $g_0$ and $g_{n-1}$, respectively.
    }
\end{figure}

For the experiments in Section~\ref{sec:AssociativeRecall},
the noisy spike times $s_0, ..., s_{n-1}$
are obtained from the nominal spike times $\breve{s}_0, ..., \breve{s}_{n-1}$
by adding zero-mean Gaussian jitter with variance $\sigma_s^2$
while maintaining the constraint that 
$s_0, ..., s_{n-1}$ are separated by at least $\taumin$.
A factor graph of the pertinent joint probability density function
is shown in \figref{eqn:JitterFG}.

Sampling from this distribution can be done by Gibbs sampling \citep{GemanGeman84} as follows.%
\sfootnote{The extension to constrained first and/or last spike, e.g., minimum starting time or periodicity requirement, is straightforward.}
For $m=1, 2, \ldots, M\gg 1$,
alternate the following two steps, 
beginning with $s_k^{(0)} = \breve s_k$ 
for all $k$:
\begin{enumerate}
\item
For even indices $k$, 
sample $s_k^{(m)}$ from a Gaussian with mean $\breve s_k$
and variance $\sigma_s^2$, truncated to the interval $(s_{k-1}^{(m-1)} + \taumin, s_{k+1}^{(m-1)} - \taumin)$.
\item
For odd indices $k$,
sample $s_k^{(m)}$ from a Gaussian with mean~$\breve s_k$
and variance $\sigma_s^2$, truncated to the interval $(s_{k-1}^{(m)} + \taumin, s_{k+1}^{(m)} - \taumin)$.
\end{enumerate}
Return $s_k=s_k^{(M)}$ for all $k$.

As always with Gibbs sampling, it is difficult to know how large $M$ needs to be.
The experiments in Section~\ref{sec:AssociativeRecall} were done with $M=1000$.

\bibliographystyle{APA}

\end{document}